\documentclass[10pt,letterpaper]{article}
\usepackage[top=0.85in,left=2.75in,footskip=0.75in]{geometry}

\usepackage{amsmath,amssymb}

\usepackage{changepage}

\usepackage{textcomp,marvosym}

\usepackage{cite}

\usepackage{nameref,hyperref}

\usepackage[right]{lineno}

\usepackage[nopatch=eqnum]{microtype}
\DisableLigatures[f]{encoding = *, family = * }

\usepackage{array}

\usepackage[utf8]{inputenc}
\usepackage{graphicx}
\usepackage{mathtools}
\usepackage{array}
\usepackage{xpatch}
\usepackage{xspace}

\usepackage{booktabs}
\newcommand*{\Pm}{\textit{Permuted}\xspace}
\newcommand*{\Rm}{\textit{Rotated}\xspace}

\usepackage{setspace}
\newcolumntype{M}[1]{>{\centering\arraybackslash}m{#1}}
\usepackage[dvipsnames]{xcolor}
\usepackage{makecell, multirow}
\usepackage[skip=1ex]{caption}
\usepackage{xargs}                      % Use more than one optional parameter in a new commands
\usepackage[colorinlistoftodos,prependcaption,textsize=tiny]{todonotes}
\newcommandx{\unsure}[2][1=]{\todo[linecolor=red,backgroundcolor=red!25,bordercolor=red,#1]{#2}}
\newcommandx{\change}[2][1=]{\todo[linecolor=blue,backgroundcolor=blue!25,bordercolor=blue,#1]{#2}}
\newcommandx{\info}[2][1=]{\todo[linecolor=OliveGreen,backgroundcolor=OliveGreen!25,bordercolor=OliveGreen,#1]{#2}}
\newcommandx{\improvement}[2][1=]{\todo[linecolor=Plum,backgroundcolor=Plum!25,bordercolor=Plum,#1]{#2}}
\newcommandx{\thiswillnotshow}[2][1=]{\todo[disable,#1]{#2}}
\usepackage{algorithm}
\usepackage[toc,page]{appendix}
\usepackage[noend]{algpseudocode}

\usepackage{xcolor}
\usepackage{amssymb} % or \usepackage{amsfonts}
\usepackage{makecell}
\usepackage{comment}

\algrenewcommand{\algorithmiccomment}[1]{\hfill// #1}

\newcolumntype{+}{!{\vrule width 2pt}}

\newlength\savedwidth

\raggedright
\usepackage[aboveskip=1pt,labelfont=bf,labelsep=period,justification=raggedright,singlelinecheck=off]{caption}

\usepackage{lastpage,fancyhdr,graphicx}
\usepackage{epstopdf}
\fancyheadoffset[L]{2.25in}
\fancyfootoffset[L]{2.25in}
\makeatletter
\def\BState{\State\hskip-\ALG@thistlm}
\makeatother
\makeatletter
\xpatchcmd{\titlepage}{\@restonecolfalse\newpage}{\@restonecolfalse}{}{}
\xpatchcmd{\endtitlepage}{\if@restonecol\twocolumn \else \newpage \fi}{\if@restonecol\twocolumn \else  \fi}{\typeout{success}}{\typeout{fail}}
\makeatother

\usepackage{minitoc}

\newcommand{\greentick}{\textcolor{ForestGreen}{\checkmark}}
\newcommand{\redcross}{\textcolor{red}{$\times$}}
\newcommand{\secbest}[1]{\textcolor{gray}{\bf #1}}

\begin{document}

\vspace*{0.2in}

\begin{flushleft}
{\Large
\textbf\newline{Context selectivity with dynamic availability enables lifelong continual learning} % Please use "sentence case" for title and headings (capitalize only the first word in a title (or heading), the first word in a subtitle (or subheading), and any proper nouns).
}
\newline
\\
Martin L.L.R. Barry\textsuperscript{1},
Wulfram Gerstner\textsuperscript{1},
Guillaume Bellec\textsuperscript{1,\Letter}
\\
\bigskip
\textbf{1} Department of Life Sciences, Department of Computer Sciences \\
École Polytechnique Fédérale de Lausanne (EPFL), Switzerland
\\
\bigskip
\Letter ~guillaume.bellec@epfl.ch
\end{flushleft}

%\blfootnote{Preprint; Under review}
\doparttoc % Tell to minitoc to generate a toc for the parts
\faketableofcontents % Run a fake tableofcontents command for the partocs
% \doublespacing
\section*{Abstract}
     "You never forget how to ride a bike", -- but how is that possible? The brain is able to learn complex skills, stop the practice for years, learn other skills in between, and still retrieve the original knowledge when necessary. The mechanisms of this capability, referred to as lifelong learning (or continual learning, CL),  are unknown. We suggest a bio-plausible meta-plasticity rule building on classical work in CL which we summarize in two principles: (i) neurons are context selective, and (ii) a local availability variable partially freezes the plasticity if the neuron was relevant for previous tasks. In a new neuro-centric formalization of these principles, we suggest that neuron selectivity and neuron-wide consolidation is a simple and viable meta-plasticity hypothesis to enable CL in the brain. In simulation, this simple model balances forgetting and consolidation leading to better transfer learning than contemporary CL algorithms on image recognition and natural language processing CL benchmarks.
\section*{Introduction}

While Artificial Neural Networks (ANNs) have made impressive advances in various computational tasks, their approach to continual learning (CL) differs from that of the human brain. Traditional ANNs, encompassing architectures like multi-layer perceptrons \cite{Rumelhart86}, Long-Short-Term Memory \cite{Hochreiter1997long}, Convolutional Neural Networks \cite{lecun1998gradient}, and Transformers \cite{vaswani2017attention}, face the challenge of integrating new knowledge without inadvertently disrupting prior information—a phenomenon termed "catastrophic forgetting" \cite{goodfellow2013empirical,kirkpatrick2017overcoming,323080} which is a manifestation of the classic stability-plasticity dilemma in computational neuroscience \cite{Carpenter88}. The human neural system, however, has a remarkable ability to assimilate new experiences without completely overwriting old ones, while generalizing previous knowledge to new problems (forward transfer) \cite{dienes1997transfer} and updating prior knowledge when learning a new, but related task (backward transfer). 
The difference in performance between ANNs and the human brain has attracted the attention both of computational neuroscientists working on synaptic learning theory 
\cite{zenke2017continual,kaplanis2018continual} as well as machine learning experts interested in biological inspirations 
\cite{goodfellow2013empirical,kirkpatrick2017overcoming,serra2018overcoming,de2021continual}.  

Cortical neurons are selective to sensory stimuli \cite{gross1972visual,olshausen1996emergence} as well as more abstract concepts \cite{quiroga2005invariant,kreiman2000category}. Populations of neurons encode abstract rules or contexts  \cite{wallis2001single,Rigotti13}, most likely implemented by modulation of their gain function \cite{Andersen85,Andersen90,van1995influence,Anderson03,Salinas96,barbosa2023early}. We conjecture that the mechanism by which humans achieve continual lifelong learning is rooted in context-specific modulation of neuronal activity \cite{Salinas04} and metaplasticity \cite{Abraham08} of synaptic connections. At a cognitive level, our hypothesis can be summarized as follows: When we encounter a new learning challenge (say we learn a new foreign language) a context-selective modulation of cortical neurons leads to partial specialization of neurons for the specific context; once selective neurons learn through synaptic plasticity (we learn this language sufficiently well), the plasticity of these neurons is then partially frozen (the new language will never be forgotten) even when the next learning task is encountered (another foreign language).

%In this paper, we propose fundamental principles of CL to formalize the above high-level hypothesis. This leads to the proposition of a metaplasticity rule for CL that we test functionally in artificial neural networks and which also yields testable predictions for real cortical circuits.

To derive a functional and plausible mechanism for CL we identified two principles that summarize the essence of existing CL models \cite{kirkpatrick2017overcoming,zenke2017continual,de2021continual,ramesh2022model,ke2021achieving,sun2020lamol,PARISI201954,NIPS2017_0efbe980,Robins1995CatastrophicFR,masse2018alleviating,flesch2023modelling} (see Discussion for an exhaustive review): (Principle 1) Neurons are {\bf gate}d to induce context selectivity, and (Principle 2) the parameters can be frozen by {\bf O}bstruction of {\bf N}eural parameter updates to prevent overriding important previous knowledge. In this paper, we suggest a  re-formalization of Principles 1 and 2 termed "GateON" which is grounded in a mathematical understanding of CL. Despite the apparent conceptual simplicity of GateON, it is less subject to forgetting and achieves better transfer learning in comparison with previous CL algorithms on established CL benchmarks such as permuted MNIST \cite{kirkpatrick2017overcoming,zenke2017continual}, split CIFAR-100 and natural language processing (NLP) benchmarks with a pre-trained BERT model \cite{devlin-etal-2019-bert}. On most metrics, GateON ranks better than algorithms involving significantly more complex mechanisms like a perfect replay of previous data samples \cite{de2021continual,PARISI201954,NIPS2017_0efbe980,Robins1995CatastrophicFR,ramesh2022model,ke2021achieving,sun2020lamol}. Additionally, we argue that GateON enables two unique and new contributions:

First, not only does the availability variable freeze parameters that are relevant to previous tasks, but it can also un-freeze parameters when resources are necessary. Hence, it balances automatically between forgetting and consolidation.
Un-freezing parameters is shown to be decisive for large task families of 100 tasks.

The second novelty is a new computational understanding of biology: we describe a \emph{neuro-centric} formalization of Principles 1 and 2, so-called n-GateON. Rather than freezing synaptic weights individually, all the neuron parameters of a neuron are freezing simultaneously. We show in particular a principled and efficient approximation of n-GateON where the recent neuron activation is directly freezing the neuron's parameter updates. Hence this simple metaplasticity hypothesis does not require non-local task-specific information like loss gradients.
This demonstration is an attempt to exhibit functional models of CL that are simple enough to be falsifiable in experimental neuroscience. Concretely the resulting mechanism could explain some features of real neurons:
the neural selectivity modulated by contextual information as widely observed in the brain \cite{Andersen85,Andersen90,van1995influence,Anderson03,Salinas96,Salinas04,barbosa2023early} is explained but the formalization of Principle 1 in GateON;  and Principle 2 in n-GateON formalizes the hypothesis that neuron-specific dynamic variables integrate the activation history and control the availability for future plastic changes which links to metaplasticity \cite{Abraham08,el2012stable} and synaptic consolidation \cite{Fusi2005-fk,Zenke2014,Ziegler15,Benna2016-cf}.
On a higher level, given the high importance of the coordination of principles 1 and 2 in our simulations, we speculate that neuron selectivity and neuron-wide consolidation might have evolved together to enable CL in the brain. 

\section{Results}
%Inspired by the literature of CL from machine learning and computational neuroscience, we describe in section \ref{sec:theory} a mathematical model of CL. This model relies on two fundamental principles: (1) gated context selectivity, and (2) obstruction of the neural parameter updates when the parameters are relevant for previous tasks. For this reason, we call it the 'GateON' principles, and we then describe in section \ref{sec:plausible} the GateON as a simple model of synaptic plasticity that could therefore support lifelong learning in biologically plausible 

Our results are presented in the following order: We first describe the mathematical principles that we identified to be the foundation of efficient CL (Section \ref{sec:theory}). Then we derive a biologically plausible implementation of this model which illustrates how neural selectivity can be the pillar of a metaplasticity rule for CL (Section \ref{sec:bioplausible}). The quantitative results about the performance of GateON on machine learning benchmarks are reported in Section \ref{sec:simulation}.

\subsection{A normative theory for continual learning}
\label{sec:theory}
In order to study CL, we follow an established paradigm \cite{kirkpatrick2017overcoming,zenke2017continual} and use a sequence of $K$ supervised tasks $T_1, T_2, \dots T_K$, where each task is also referred to as a 'context'.
The performance of the network on task $T_k$ is quantified by the loss $\mathcal{L}^k$ averaged over all data points in that task. 
The same task  $T_k$ is used for an unknown time before a switch to another task occurs. A single time point $t$ corresponds to the presentation of a single data point (or a minibatch of data points) at the input layer of the network, followed by network processing, and calculation of the loss based on the network output. The instantaneous loss for a single data point (or a minibatch of data points) is $\mathcal{L}^t$. 

\paragraph{Principle 1: Gated context selectivity.}
To define a model of contextual selectivity in neural network models, we assume that each network unit is gated by a context-dependent sigmoidal gate. We describe here the case of a  feedforward network comprised of $L$ layers with $N$ neurons per layer (the extension to Transformers is highlighed in section \ref{sec:simulation}  and to Convolutional Neural Networks (CNNs)  in {\bf{Methods}} \ref{matmet:CNNGateON}). The activity $x_i^l$ of a neuron $i$ in layer $l\ge 1$ is determined by a multiplicative gain function \cite{Salinas96,Salinas04} 
with gating factor $g_i^l$ and activation function $f$:
\begin{equation}\label{eq:multiplicative}
x^{l}_i = g^{l}_i \cdot f \left( \sum_j w^{l}_{ij}x^{(l-1)}_j \right), \quad \text{where } g^{l}_i = \sigma \left( v^{l}_{ik}\right).
\end{equation}
Here, $w_{ij}^{l}$ is the connection weight from neuron $j$ in layer $l-1$ to neuron $i$ in layer $l$, while $v^{l}_{ik}$ represents the context-specific 'gating weights' that control neuron $i$ in layer $l$. The index $k$ which determines which context is active is considered to be given as input in the CL literature \cite{serra2018overcoming,iyer2022avoiding} and we follow this assumption in this theoretical section. Biases (thresholds) are treated as additional weights for simplicity. The function $\sigma$ is chosen as a rectified hyperbolic tangent to ensure that the range of $g_i^l(t)$ is confined between 0 and 1. The network input is represented by $x^{(0)}_i$ and the output by $y_i = x^{L}_i$. The gating weights are initialized randomly and learned with normal gradient descent (Principle 2 does not apply to these parameters). In effect, this gating selects a distributed subset of neurons that actively participate in the information processing for a context $k$ whereas other neurons that participate in other contexts are, at least partially, turned off. Mechanisms like lateral inhibition \cite{Barry2022,masse2018alleviating} or divisive normalization \cite{Heeger96, Carandini99} could implement a similar gating in a biophysical network model. 

\paragraph{Extension of Principle 1 to unidentified context switches.}
In general, a context switch is not marked: for example, toddlers in a multi-lingual environment are expected to infer by themselves change points between two spoken languages. To illustrate that the GateON Principles also apply when the context identities (i.e. the task index) are not provided, we also studied a variation of  Principle 1 in which the task index $k$ at each time step is inferred by detecting a sudden increase in the loss $\mathcal{L}^t$. Simply put, this alternative model indicates a change point each time a moving average of $\mathcal{L}^t$ crosses some threshold (see {\bf Methods \ref{matmet:taskinference}} for a precise formalization, this simple model is also inspired by computational neuroscience models \cite{Barry2022,Liakoni21}). Importantly, while with this simple model variation of contexts are inferred without supervision, no additional changes in the GateON theory are required (see section  \ref{sec:simulation}). Other context detectors have been studied in more detail \cite{caccia2020online,heald2021contextual} and could have been used in conjunction with GateON. To yield a fair comparison with machine learning algorithms, we use the standard CL paradigm where the task index $k$ is given directly. 

\paragraph{Principle 2: Gradient Obstruction - freezing and unfreezing of plasticity.}
Additionally to the context-selectivity, a functional CL model requires a second ingredient: in our conjecture when neurons are specialized for a task, their parameters should freeze to avoid later overwriting of acquired knowledge.
Assuming that during context $k$ plasticity implements gradient descent of the loss $\mathcal{L}^k$, we formalize below how, and when, the gradient descent updates need to be obstructed (Principle 2).
To make the relationship with the machine learning literature explicit, we first formulate Principle 2 in a parameter-centric view where the update of a parameter $\theta$ is obstructed.

\paragraph{The parametric view.}
The parametric view of GateON (p-GateON) is inspired by previous CL models \cite{de2021continual,rusu2016progressive, NEURIPS2018_cee63112} which maintain stable weight vectors in the feedforward processing path to mitigate catastrophic forgetting. 
Yet, it is possible to build examples where naively preserving previously optimized weights can impede training for subsequent tasks - so that a partial unfreezing of previously learned weights is preferable. To navigate the challenges, the novel theory of p-GateON is formalized using an 'adaptive learning rate' that flexibly modulates gradient updates of individual parameters.
More precisely, updates of a parameter $\theta$ are proportional to the gradients with a learning rate that is controlled by the 'availability' $A_\theta$:
\begin{equation}
\Delta \theta = \eta ~ A_{\theta} \frac{\partial \mathcal{L}^t}{\partial  \theta}
\label{eq:gradientspecific}
\end{equation}
%\begin{split}
%\Delta w^{l}_{ij}=
%\begin{cases}
%    \eta \, A_{w^{l}_{ij}} \frac{\partial \mathcal{L}^t}{\partial  w^{l,}_{ij}},& \text{if p-GateON} \\
%    \eta \,  A_i^l \,  \frac{\partial \mathcal{L}^t}{\partial  w^{l,}_{ij}},           & \text{if n-GateON}
%\end{cases}
%\end{split}
%\label{eq:gradientspecific}
%\end{equation}
where $\eta$ is the nominal learning rate (SGD, Adam, \dots).  The availability $A_\theta$ is bounded between zero and one and will be defined in the next paragraph. As shown in Eq. \eqref{eq:gradientspecific}, the smaller the availability $A_{\theta}^l$, the larger the obstruction of the gradient. In otherwords, our method can freeze each parameter individually.
Note that only the feedforward weights $w^{l}_{ij}$ are modulated by the availabilities, the contextual weights $v^{l}_{ik}$ are optimized with standard gradient descent.

\paragraph{Definition of the availability.}
We define the 'availability' $A_{\theta}$ to track whether the parameter $\theta$ has been relevant for recent tasks. High relevance leads to low availability for future tasks so further parameter updates are obstructed. Mathematically $A_\theta$ is an integrator of the 'normalized relevance' $\mu^{norm}_{\theta}$ which measures the causal impact of $\theta$ of the task performance $\mathcal{L}^k$ for previous tasks. A major innovation here is the definition of the 'normalized relevance' $\mu^{norm}_{\theta}$ given in the next paragraph. The availability  integrates the quantity $\mu^{norm}_{\theta}$ over time using the formula:
\begin{equation}
A_{\theta}(t+1) = [A_{\theta}(t)(1 - \eta_A(\mu^{norm}_{\theta} - \epsilon))]_0^1,
\label{eq:availabilty}
\end{equation}
 where $\eta_A$ and $\epsilon$ are non-negative constants. The clip function denoted $[x]_0^1$ is the identity function inside $[0,1]$, takes a value of 1 for $x>1$ and a value of 0 for $x<0$. By construction, Eq. \eqref{eq:availabilty} has two stable fixed points for $A_\theta=0$ and $A_\theta=1$ and $\epsilon$ is a threshold controlling the convergence of this equation: the availability decays exponentially to zero as soon as $\mu^{norm}_{\theta}>\epsilon$, or it increases exponentially fast to $1$ if $\mu^{norm}_{\theta} <\epsilon$.
 To summarize, if a parameter is relevant $(\mu^{norm}_{\theta}>\epsilon)$ the availability $A_{\theta}$ converges to $0$ which freezes the parameter $\theta$ by obstructing subsequent parameter updates.
When a parameter later becomes irrelevant for another task, the availability increases again slowly leading to unfreezing of the parameter. The relation between the time scales of freezing and unfreezing is discussed in {\bf Supplementary \ref{matmet:fixed-point}}.
%The reset time necessary of the availability can be estimated which is useful to choose the hyperparameter $\eta_A$ in simulations; see {\bf Methods \ref{matmet:fixed-point}}.
The hyperparameter $\epsilon$ is important to balance the trade-off between preserving representations of previous tasks and maintaining the flexibility of the network for future tasks. The effect of $\epsilon$ will be studied in simulations.

\paragraph{Algorithmic relevance estimation.}
An important component of the theory is to measure $\mu_{\theta}^{norm}$ to quantify the causal impact of $\theta$ on the task performance.
Ideally the causal impact of the neuron $i$ on the loss $\mathcal{L}^t$ would be defined as the difference of performance between the current loss  $ \mathcal{L}^t$ and the hypothetical loss if the parameter $\theta$ would be removed in the same setting:
\begin{equation}
\mu_{\theta} = (\mathcal{L}_{|\theta=0}^t - \mathcal{L}^t)^2,
\label{eq:causal_effect}
\end{equation}
where $\mathcal{L}_{|\theta=0}^t$ denotes the loss under the assumption that $\theta=0$ for the same data-point.
Intuitively, a parameter is relevant if the loss changes by a large amount when it is removed.  The evaluation of the relevance in \eqref{eq:causal_effect} is computationally expensive, as it would require evaluating the loss function once per parameter.
 We therefore approximate it by its first-order Taylor expansion around the current set of parameters, i.e., $\mathcal{L}_{|\theta=0}^t = \mathcal{L}^t - \frac{\partial\mathcal{L}^t}{\partial \theta}\, \theta$. Eq. (\ref{eq:causal_effect}) then becomes:
\begin{equation}\label{eq:p-mu}
    \mu_{\theta} \approx \left( \frac{\partial\mathcal{L}^t}{\partial \theta}\theta\right)^2.
\end{equation} 
Note that $\frac{\partial\mathcal{L}^t}{\partial \theta}\theta$ can be calculated efficiently since it involves the same gradient with respect to parameters that is also used in BackProp.
Finally, we normalize $\mu_{\theta}$ so that GateON is insensitive to the scaling of the loss function $\mathcal{L}^k$ and the parameters $\eta_A$ and $\epsilon$ do not have to be fine-tuned for every CL problem. To do so, we use a softmax normalization
\begin{equation}
\begin{split}
    \mu^{norm}_{\theta} =& N^2 \frac{\mu_{\theta}}{\sum_{\theta'} \mu_{\theta'}},\\
\end{split}
\label{eq:mujN}
\end{equation}
where $N^2$ is the number of parameters $w_{ij}^l$ in the layer $l$ ($N$ is the number of neurons). As a consequence $0\le\mu_\theta^{norm}\le N^2$ and the availability dynamics which is insensitive to a re-scaling of the loss because it it is canceled by softmax in Eq. \eqref{eq:mujN}. Noting that value $\mu^{norm}_{\theta}=1$ corresponds to a uniform softmax distribution, so the multiplication by $N^2$ makes sure that a threshold value $\epsilon=1$ automatically balances between freezing or un-freezing parameters: When the availability variables change, there is at least one parameter is deemed irrelevant and un-freezes ($\mu^{norm}_{\theta} < 1$) and one relevant parameter that freezes $\mu^{norm}_{\theta} > 1$. In other words, GateON with parameter $\epsilon=1$ is designed to balance between forgetting and consolidation, in simulations we also tested $\epsilon=0.5$ which favors slightly consolidation over forgetting.

\subsection{Bio-plausible implementation of GateON}
\label{sec:bioplausible}

We describe now an extension of the GateON theory and justify why we believe this alternative implementation is more compatible with existing biological data and mechanisms.

\paragraph{The selectivity of cortical neurons to experimental context supports Principle 1.}
To begin with, we view Principle 1 (the gated context selectivity, Eq. \ref{eq:multiplicative}) as a simple model of the neural selectivity that is widely observed in the brain. Classically in visual cortices, it is well known that neurons are highly selective to visual stimuli \cite{gross1972visual} such that the population of selective neurons builds together a sparse and distributed code of visual stimuli \cite{olshausen1996emergence} or more abstract concepts  \cite{quiroga2005invariant}.  Importantly, contextual rules are represented in non-sensory cortical areas such as prefrontal cortex: when training a monkey on multiple visual tasks, neurons develop distributed feature representations including selectivity to task rules \cite{wallis2001single,Rigotti13}. 
We see our gated context selectivity model of Eq. \eqref{eq:multiplicative} as a minimal model of these general observations in line with earlier models of multiplicative gain function modulation \cite{Salinas96,Salinas04}.

\paragraph{A plausible neuro-centric implementation of Principle 2.}
Given the context selectivity of cortical neurons, we conjectured that a hidden biological mechanism capable of tracking the context relevance over long time scales is likely to be found at the neuronal level rather than the synaptic level. Under this assumption, we look for a neuro-centric implementation of Principle 2 where the availability variable is related to neuronal variables (such as firing rate or membrane potential) instead of auxiliary synaptic variables. Using the same normative theory that we used to derive p-GateON, we now derive n-GateON as an analogous, but neuro-centric, model.

\begin{figure}[t!]
    \centering
    \includegraphics[width=1\linewidth]{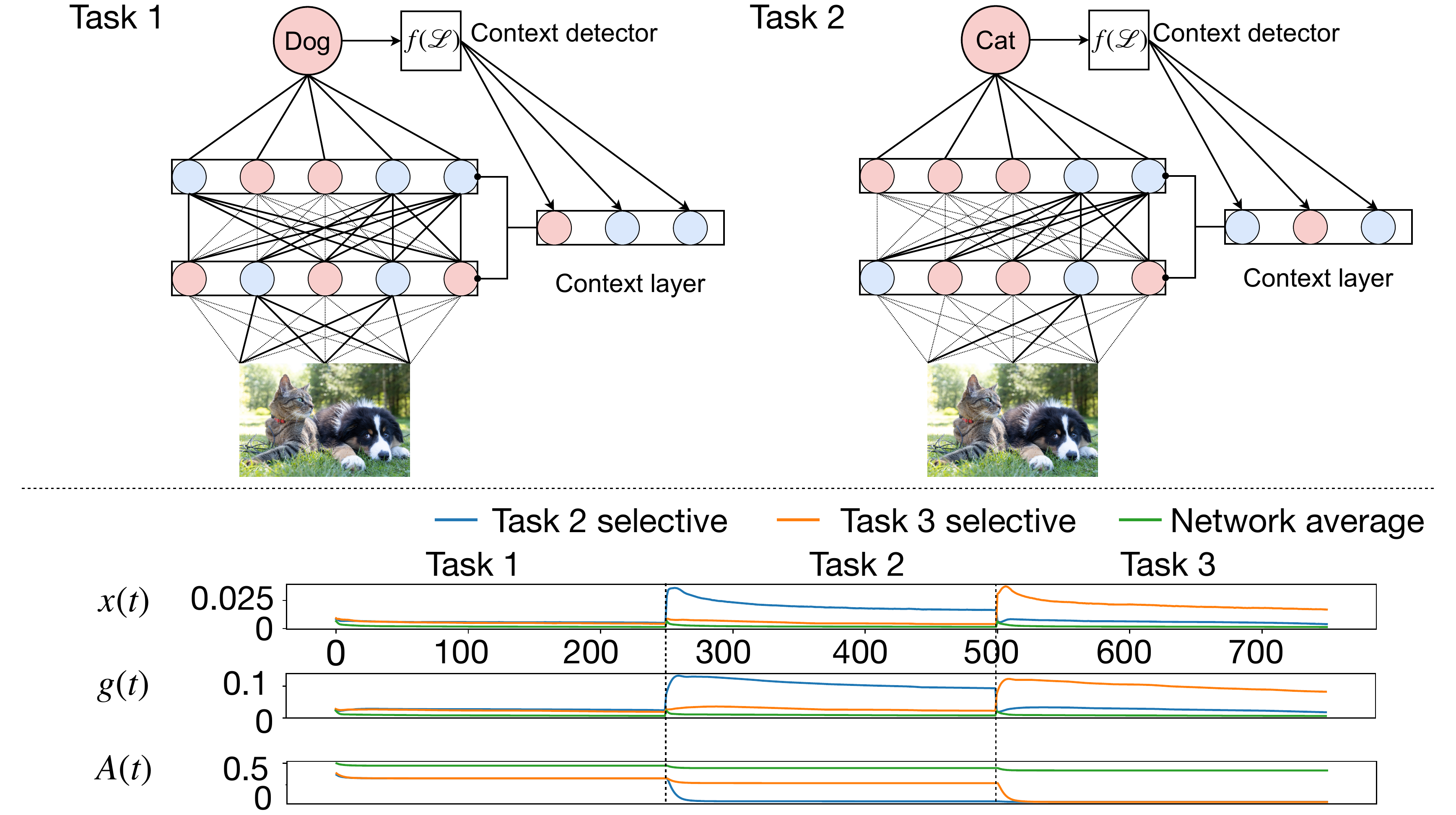}
    \caption[GateON schematic]{{\bf{Neural selectivity and learning availability. Top Panel}}: Illustration of the GateON selectivity mechanism across two sequential tasks. Consider the scenario wherein task 1 classifies the animal on the left side of an image, while task 2 classifies entities on the right. In the neural network, the context layer silences the neurons in blue and leaves the red ones active. Connections illustrated as dashed lines represent those with diminished availability by the end of the task. {\bf{Bottom Panel}}: A representative simulation of n-GateON spanning a sequence of three tasks using permuted MNIST. We display the mean neural activity of task-selective neurons ({\em{x}}, top), the gating state of task-selective neurons ({\em{g}}, middle), and availability ({\em{A}}) for 250 mini-batch presentations. The blue and yellow trajectories represent averages across those neurons with activity above some threshold during tasks 2 or 3.
    %Neurons demonstrating selectivity for tasks 2 or 3 manifest heightened gating values specific to their respective tasks, surpassing the overall network average (represented in green). As tasks progress, the availability of task-specific neurons gradually decays toward zero (parameter $\epsilon =0$). It's important to note that while the average availability for a majority of neurons remains elevated (shown in green), the availability for neurons exclusive to task 2 is retained at a diminished level throughout the presentation of task 3, underscoring that only a minority of neurons are deemed pertinent for the task at hand.
    }
    \label{fig:GateONscheme}
\end{figure}

We consider that each neuron has an availability variable $A_i^l$ which obstructs the plasticity of all the parameters of neuron $i$ in layer $l$. Similarly, as previously, this availability $A_i^l$ should integrate the causal impact of the neuron $i$ on the loss $\mathcal{L}^k$. Hence in this neuro-centric perspective Principle 2 is formalized by the two equations:
\begin{equation}
\Delta w^{l}_{ij}= \eta \,  A_i^l \,  \frac{\partial \mathcal{L}^t}{\partial  w^{l,}_{ij}}
\quad \text{ and } \quad
\mu_{i}^l  \approx  \left(\frac{\partial\mathcal{L}^t }{\partial x^l_i}x^l_i \right)^2,
\label{eq:n-mu}
\end{equation}
where $x_i^l$ is the activity of neuron $i$ is layer $l$. The availability is updated otherwise as in Eq. \eqref{eq:availabilty} where we normalize $\mu_{i}^l$ in the range $[0, N]$ instead of $[0, N^2]$ since we have $N$ neurons instead of $N^2$ parameters. Having significantly fewer availability variables in this model will have advantages and disadvantages that we analyze in the simulations in Section \ref{sec:simulation}.
Figure \ref{fig:GateONscheme} bottom illustrates an example of the availability of task-selective neurons across three tasks.
%We discuss the network inference of the context in {\bf{additional figure}}  \ref{matmet:fig:taskinference}. 

\paragraph{Further simplification of the relevance computation?}
As it stands, we view the formalization of n-GateON as a convincing biologically plausible realization of Principles 1 and 2. The model n-GateON will be implemented as described above in all simulations unless explicitly stated otherwise. However, it may argued that the plausibility of the model is incomplete because the derivative $\frac{\partial\mathcal{L}^t }{\partial x^l_i}$ in Eq. \eqref{eq:n-mu} requires a biologically plausible implementation of Backprop which, by itself,  remains an open problem in the field. 
This issue is that a mechanism distinct and equally complex as the network itself is necessary to compute the backward pass but is not found in the brain \cite{Richards19}. It gives rise to biologically plausible approximations of back-prop \cite{Payeur21,bellec2020solution,golkar2022, Illing2021}, and importantly here,
 most of them are compatible with our model.  To compute the gradient $\frac{\partial\mathcal{L}^t }{\partial \theta}$ with local variables or bio-plausible top-down signals, the hardest part of these theories is to provide a plausible computation of the gradient $\frac{\partial\mathcal{L}^t }{\partial x^l_i}$. Since $\mu_i^l$ relies on the exact same term, any combination GateON with such a model is viable as it relies on the same term. 
%I DID NOT GET THE MEANING OF THE FOLLOWING.
%Going further, the GateON hypothesis provides measurable quantitative predictions if one combines our theory with learning rules validated experimentally: for instance, in a simple reinforcement learning model $\frac{\partial\mathcal{L}^t }{\partial x^l_i}$ can be viewed as reward prediction error communicated through dopamine \cite{bellec2020solution, fremaux2016neuromodulated}. So Eq. \ref{eq:mujN} already provides a precise and measurable experimental prediction of our model. 

\begin{figure}[t!]
    \centering
    \includegraphics[width=1\linewidth]{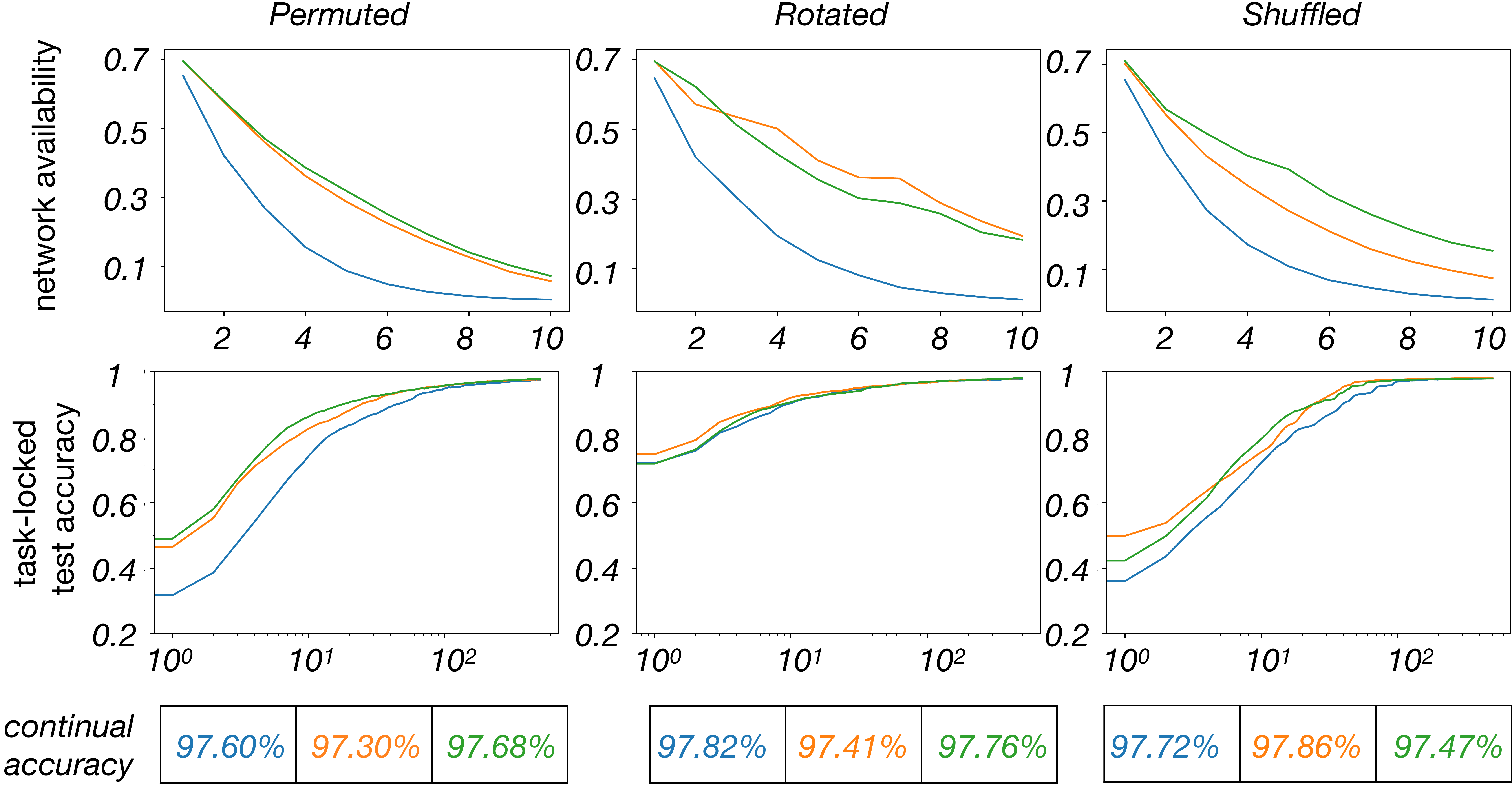}
    \caption{{\bf Effect of bio-plausible simplification of the relevance measure in n-GateON}. Top: The average availability $\langle A_i^l\rangle$ of neurons (evaluated at times $t_k$ for tasks $1\le k \le 10$) decreases over 10 tasks for MNIST CL problems {\em Permuted} (left), {\em Rotated} (middle) and {\em Shuffled} MNIST (right) using n-GateON with $\epsilon=0$.
    The blue curve shows results with the {algorithmic relevance of Eq. (\ref{eq:n-mu})}, the green one with the simplified bioplausible relevance of Eq. (\ref{eq-mu8}) % {\color{ForestGreen} the plausible layer-wise approximation $\mu_i^l = {x_i^l }^2$} and 
    and the orange one shows an intermediate bioplausible variant for comparison (see {\bf Methods \ref{matmet:layer-wise-derivation}}).
    %With the biologically plausible relevance measure,  the availability of neurons decreases faster which implies that after 10 tasks more neurons show 'frozen' plasticity than with the other two relevance definitions. 
    %Middle:  Task-locked test accuracy of task $T_k$  as a function of training time $t-t_{k-1} $ for 500 training steps per task, averaged over tasks (see Methods \ref{matmet:tasklocked}). 
    Middle and Bottom: With all three relevance definitions, n-GateON approaches a similar value of a task-locked accuracy  (middle) and continual accuracy (bottom) above $97\%$ after 500 training steps per task.
    %Bottom: Similarly, the continual accuracy across 10 tasks is also above $97\%$.
    }
    %meaning that the CL capabilities of the network stays the same. In practice, the approximated relevance mapping only impacts the saturation of the network but not directly its CL properties.}
    \label{matmet:fig:comparisonava}
\end{figure}

We found however that an alternative derivation of the causal influence of the neuron $i$ on the rest of the computation is also possible. We derive in {\bf Methods \ref{matmet:layer-wise-derivation}} a formula for $\mu_i^l$ to evaluate the effect of the neuron $i$ on the next layers rather than the loss itself. It turns out that this derivation is proportional to 
\begin{equation}\label{eq-mu8}
    \mu_i^l = ({ x_i^l }) ^2 
\end{equation}
 which yields a simplified relevance definition that is much easier to compute locally without feedback from the downstream network.  Fig. \ref{matmet:fig:comparisonava} shows that the approximation of Eq. \eqref{eq-mu8} is functional although slightly less efficient than the gradient-based definition of n-GateON in Eq. \eqref{eq:n-mu}.

\subsection{Framework for Simulation Results}
\label{sec:simulation}
Before reporting the simulation results for p-GateON and n-GateON on CL benchmarks, we introduce multiple metrics to quantify the success of a CL model.

Tasks $T_1,T_2, ..., T_K$ are presented one after the other and training stops when the last task $T_K$ has finished.
Let $t_0$ denote the time when the first task starts and
$t_k$ the time point when the last data for task $T_k$ is given. 
In CL several aspects of the network robustness and performance can be tested.
Classically, standard neural networks trained with gradient descent are subject to catastrophic {\bf forgetting}, which is visible when the performance drops drastically for earlier tasks $T_k$ when new tasks $T_{k'}$ with $k' > k$ are learned.
Another problem emerges specifically with CL algorithms that aim to 'freeze' parameters in order to hold the knowledge of previous tasks: if all the model parameters are frozen after $k$ tasks, we say that the model is {\bf saturated} which undermines its performance on any future task $k'>k$.

The last question is whether the model is capable of re-using previous knowledge for future tasks: the accuracy on task $T_k'$ is better after learning task $k$ than if task $k'$ is learned in isolation. If so, this is a sign of {\bf forward transfer}.
Reversely, learning a later task $T_{k'}$ can improve the accuracy on task $T_{k}$ which is a sign of {\bf backward transfer}.
In practice, we employ four quantitative measures in Methods \ref{matmet:measures} to quantify these high-level aspects:

(i) The {\em{immediate test accuracy $A^k_{cc}$}}, measures the test accuracy on a single task $T_k$  immediately after training this task, i.e., after the update step at time $t_k$. It measures {\bf saturation}. 

(ii) The {\em{continual accuracy $A_{cc}^{cont,k}$}}, is computed by testing the accuracy on task $T_k$ after training all later tasks $T_{k'}$ with $k'>k$. The quantity is averaged for all $k'>k$. The {\em continual accuracy} is a combined measurement of {\bf saturation} and {\bf forgetting}.

(iii) The {\em{accuracy deviation $\Delta A^k_{cc}$}}, computes the relative difference between the immediate test accuracy on task $T_k$ and the accuracy of the same network trained only on $T_k$ ($\Delta A^k_{cc}>0$ for $k\ge 2$ implies {\bf forward transfer}).

(iv) The {\em{forgetting rate \textit{FR}$^k$}}, is the difference between the immediate test accuracy and the continual accuracy. A positive \textit{FR}$^k$ means that the model has {\bf forgotten} task $k$, while a negative \textit{FR}$^k$ implies that the previously trained task $T_k$ increases its test accuracy during training of later which is an evidence for {\bf backward transfer}. 

For all four measures, we omit the index $k$ when the measure is averaging over all task $k$. For instance, the forgetting rate \textit{FR} is the averaged of all \textit{FR}$^k$.

\paragraph{Model comparison on established image CL problems}
 To investigate the properties of the GateON method we apply it to four standard CL problems \cite{goodfellow2013empirical,srivastava2013compete,lopez2017gradient} derived from the MNIST dataset. For each of the four problems, the network receives the image pixels as inputs and gives a digit label as output. 
 By changing the input-output convention in different ways in the four different models, we can alternatively measure the robustness to: random input pixel permutations (\textit{permuted MNIST}), structured input changes via image rotation (\textit{rotated MNIST}), random output changes (\textit{shuffled MNIST}) and incremental addition of new label classes (\textit{split MNIST}). When it is clear from the context that we refer to \textit{permuted MNIST} we will refer to this benchmark as \textit{permuted}. The task specifications are given in {\bf Methods \ref{matmet:MNIST}}

To enable a comparison with State-of-the-Art CL methods, we tested n-GateON and p-GateON on several CL problems across $K=10$ MNIST-related tasks (Tab. \ref{tab:accMNIST10}).
We consider a network with 2000 hidden neurons per layer. We observe that GateON performs well across {\em all} three established CL MNIST problems. The continual accuracy that is reached is close to the performance of 'Isolated' models optimized for each task separately.
We emphasize in particular that GateON achieves high continual accuracy despite its conceptual simplicity. For instance, GateON does not replay samples from previous tasks as in \cite{iyer2022avoiding,ramesh2022model}. Similarly, it does not use task-dependent network outputs as in \cite{farajtabar2020orthogonal,serra2018overcoming}. While most existing CL algorithms cannot be mapped onto mechanisms as simple mechanisms as GateON (only EWC and SI are comparatively simple), GateON achieved a continual accuracy higher than all other CL models on comparable task specifications. Interestingly p-GateON and n-GateON are both reliably efficient across \textit{permuted},  \textit{rotated},  \textit{split} and \textit{shuffled MNIST} (see also Table \ref{tab:ablation}) showing that the generic GateON theory works for various types of input and output CL task variations.

For further comparison, we applied GateON to a  CL problem derived from the CIFAR 100 image dataset using a ResNet convolutional network model rather than a fully connected model. The problem, \textit{Split} CIFAR 100, contains 20 tasks constructed with object class pairs appearing incrementally.
The continual accuracy of GateON is above all replay-free models we found in the literature (Table \ref{tab:accCIFAR}). 
Interestingly, on this task, n-GateON performs better than p-GateON.
% ARE YOU SURE THAT THIS IS TRUE? : perhaps because there are many more units than parameters in convolutional neural networks. 

\begin{table}
\centering
  \begin{tabular}{|l||l|l|l|c|c|}
    \hline
     {\em Models} & 
     \textit{Permuted} & 
     \textit{Rotated} & 
     \textit{Split} & 
     {\em Replay-free} & {\em \makecell{Fixed model}} \\
    \hline
     \multicolumn{6}{c}{} \\
    \hline
     EWC\cite{kirkpatrick2017overcoming} &  96.9 &  84 & -  & \greentick & \greentick \\
     \hline 
     SI\cite{zenke2017continual} & 97.1 & 98.9 & - & \greentick & \greentick  \\
   
    \hline
    
    n-GateON 0   ({\bf ours}) & \bf{97.8} & 99.2 & {\bf{99.98}} & \greentick & \greentick  \\
    \hline
    p-GateON 0   ({\bf ours}) & 97.3 & {\bf 99.3} & {\bf{99.98}} & \greentick & \greentick \\
     \hline
    RMN\cite{kaushik2021understanding} & 97.7 & - & 99.5 & \greentick & \greentick  \\
    \hline
    Active dendrite\cite{iyer2022avoiding} & 97.2 & - & - & \greentick & \greentick  \\
    \hline
    
     \multicolumn{6}{c}{} \\
    \hline
         OGDT\cite{farajtabar2020orthogonal} & 86.4  &   88.3 & 98.8 & \greentick & \redcross  \\
           \hline
    HAT\cite{serra2018overcoming} & (98.6)  & - & 99.0 & \greentick & \redcross  \\
     \hline
    Zoo \cite{ramesh2022model} & 97.7  & ({{99.7}}) & 99.97 & \redcross & \greentick  \\
    \hline
     \multicolumn{6}{c}{} \\
    \hline
    {\em Isolated tasks} & {\em 98.0 }& {\em 99.4} & {\em99.99} &  \multicolumn{2}{c|}{}   \\
    \hline
  \end{tabular}

     \vspace{2mm}
  \caption[Results MNIST 10]{{\bf Comparison of continual learning models on MNIST.} Continual accuracy  $A_{cc}^{cont}$ (in percent) for the three classical CL-paradigms (\textit{Permuted}, \textit{Rotated}, and \textit{Split} MNIST) over 10 tasks.
  "Replay-free" means that the CL algorithm does not re-sample data from previous tasks, and "Fixed model" means that the model does not use task-specific readout layers.
  %We compare the performance of our model with other biologically inspired 'online' models, such as  EWC and SI as well as algorithmic machine learning continual learning models like RMN or HAT. %{Bio-inspired methods are indicated by $^{\mathcal{B}}$. } 
  GateON 0 refers to our model with $\epsilon=0$. 
   Isolated tasks refer to the averaged accuracy for 10 vanilla networks trained on each task separately. 
  %Algorithms references are: EWC \cite{kirkpatrick2017overcoming}, SI \cite{zenke2017continual}, RMN \cite{kaushik2021understanding}, OGDT \cite{farajtabar2020orthogonal}, HAT \cite{serra2018overcoming}, Active Dendrites \cite{iyer2022avoiding}, Zoo  \cite{ramesh2022model}.
  The \Rm value for Zoo is in parentheses because they used only 5 fixed small rotations as opposed to random ones. 
  The permutated MNIST performance of HAT is in parentheses because we suspect a difference in the task specifications given the large result margin above the isolated tasks baseline, we were unable to reproduce this result.
  \label{tab:accMNIST10}}
\end{table}

\begin{table}[t!]
\centering
  \begin{tabular}{|l||l|c|}
    \hline
     {\em Models} & \textit{Split CIFAR 100}  \\
    \hline
    % \multicolumn{2}{c}{} \\
    \hline
     EWC \cite{kirkpatrick2017overcoming} & 75.3  \\
     %\hline 
     %SI$^{\mathcal{B}}$\cite{zenke2017continual} & - & - \\
   
    \hline
    n-GateON 0   ({\bf ours}) &  {\bf 84.6} \\
    \hline
    p-GateON 0   ({\bf ours}) &  82.5 \\
     \hline
    RMN\cite{kaushik2021understanding} &  80.1 \\
    \hline
     %\multicolumn{2}{c}{} \\
    %\hline
    %     OGDT$^{\mathcal{H}}$ \cite{farajtabar2020orthogonal} & -\\
    %       \hline
    %HAT$^{\mathcal{H}}$\cite{serra2018overcoming} & - \\
    % \hline
    % \hline
    %Active dendrite$^{\mathcal{R}}$\cite{iyer2022avoiding} &  - \\
    %\hline
    %Zoo$^{\mathcal{R}}$ \cite{ramesh2022model} & 95.38\\
    %\hline
    \hline
    {\em Isolated tasks}  & {\em 85.4 }  \\
    \hline
  \end{tabular}
     \vspace{2mm}
  \caption[Results MNIST 10]{{\bf Efficiency of the bio-plausible version of n-GateON on CIFAR-100.} Continual accuracy (in percent) on the task \textit{Split} CIFAR-100 which is our hardest image CL challenge.
  %We report the performance of n-GateON with the bio-plausible relevance measurement $x_i^2$ instead of the theoretical value $\left( \frac{\partial \mathcal{L}}{\partial x_i} x_i \right)^2$. 
  }\label{tab:accCIFAR}
\end{table}

\paragraph{Ablation study with 100 MNIST tasks.} 
When the number of tasks increases, the effects of network saturation and forgetting become more critical since the tasks are competing for the limited network resources.
Therefore we considered larger problems consisting of  100 \textit{permuted}, \textit{rotated}, and \textit{shuffled} MNIST tasks. In the case of \textit{Shuffled} MNIST we increase the network size from $2000$ to $5000$ neurons per layer (see methods).

To study the importance of Principle 1 and 2 separately in the GateON theory, we compare the results of GateON with four other models: an ANN of the same size trained with vanilla gradient descent (i.e., neither Principle 1 or 2 are implemented); and two ablated models 'Gating only' and 'Obstruction only' which implement only Principle 1 or 2, but not the other one.
The vanilla gradient descent model underperforms for all tasks, a phenomenon previously observed when studying catastrophic forgetting \cite{mccloskey1989catastrophic,abraham2005memory}.
The results of Table \ref{tab:ablation} show that both gating and obstruction improve substantially the performance over the vanilla baseline, but none of them work competitively on their own. As a reference point, the Active Dendrites method \cite{iyer2022avoiding} achieved $91.2\%$ continual accuracy on the permutated task with $100$ task which is higher than the ablated models but lower than p-GateOn and n-GateON which achieve $96.5\%$ and $95.7\%$ of continual accuracy respectively.
Note that most of the other models that we reviewed previously were only tested on $10$ and not $100$ tasks, and we argue below that with an increasing number of tasks, the un-freezing mechanism like the normalization of the relevance via the threshold $\epsilon$ in GateON becomes crucial.

\begin{figure}[t!]
    \centering
    \includegraphics[width=1\linewidth]{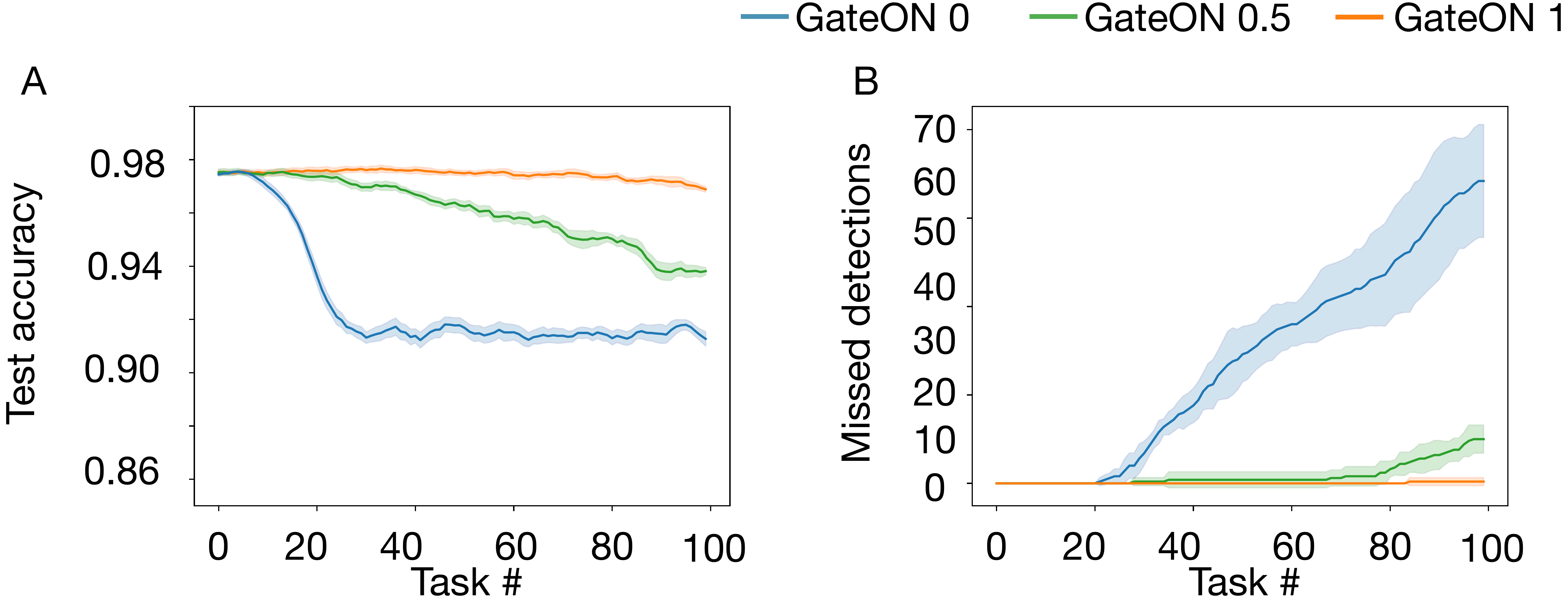}
    \caption[Leaky availability mapping and network saturation]{{\bf{The $\epsilon$ parameter affects network saturation}}. The parameter $\epsilon$ can be understood as the rate for unfreezing parameters.  {\bf{A}}: Immediate test accuracy (in percent) across tasks for \Pm on n-GateON (with task identity given, not inferred). The colors refer to different values of the parameter $\epsilon$.  For  $\epsilon = 0$ (blue), the test accuracy drops after about 10 -  20 tasks indicating that the network saturates and cannot learn new tasks.
    {\bf{B}}: Simulation where the context index $k$ in inferred. We show the fraction of missed detections of task switches for n-GateON (in percent). }
    \label{fig:capacity}
\end{figure}

\paragraph{\bf Freezing and unfreezing of parameters with the hyper-parameter $\epsilon$.}
We now show that the parameter $\epsilon$ becomes crucial to control the trade-off between network saturation and forgetting when the number of tasks is large.
Although our method with the parameter $\epsilon=0$ has achieved state-of-the-art results with $10$ MNIST tasks,  network saturation problem emerges with $100$ MNIST tasks: once all parameters are frozen the network cannot learn new tasks. This problem has not been identified in previous literature because of the small number of tasks, yet we see in table \ref{tab:ablation} that the saturation problem is most critical for unstructured tasks like \textit{Permuted} and \textit{Shuffled} MNIST. We explain below how the $\epsilon$ parameter can mitigate this problem.

First let us focus on the mathematical definition of the availability variable in Eq. \eqref{eq:availabilty}. With a choice of $\epsilon=0$, the only stable fixed point of the availability is at zero which implies that eventually all parameters will be frozen and the model saturates (it cannot learn anymore). By contrast, with $\epsilon>0$, if a parameter (or a neuron) is not relevant for the current task, it takes a value of $\mu_{\theta}^{norm}=0$, and as a consequence its availability increases exponentially fast to $1$ enabling to overwrite this parameter. It avoids complete {\em saturation} but enables {\em forgetting} of previous knowledge. When a parameter relevance is higher than the average relevance, the normalized relevance variable $\mu_{\theta}^{norm}$ goes above the value of one, so the choice $\epsilon=1$ offers a generic threshold to freeze and unfreeze depending on the relative relevance of the parameters.
In this sense, setting $\epsilon$ in this range provides an intermediate value that is designed to trade off between {\em forgetting} and {\em saturation}. 

In practice, we tested the values $\epsilon=0$, $0.5$, or $1$.  The impact on saturation and forgetting is illustrated in Fig. \ref{fig:capacity}A. The evolution of immediate test accuracy during the sequential training of 100 tasks reveals a distinct drop in accuracy after 10 tasks for n-GateON with $\epsilon=0$ due to the saturation of the model. We observe that this issue is mitigated when $\epsilon=1$ and $\epsilon=0.5$. In Fig. \ref{fig:capacity}B we show another consequence of saturation, in the setting where the task index $k$ is inferred at each sample and not given as an input. We find that a saturated model becomes incapable of inferring the task index when it gets saturated. 
The impact of $\epsilon$ on the overall continual accuracy on 100 MNIST tasks is reported in Table \ref{tab:ablation}. 

A drop in continual accuracy may reflect saturation or forgetting. However, for $\epsilon=0$ where we expect saturation, this is visible on \textit{Permuted} MNIST with n-GateON only $73.6\%$ of continual accuracy due to saturation but this performance increases to $95.7\%$ with $\epsilon=1$. Interestingly, we also observe that the saturation problem appears to be stronger with n-GateON than p-GateON.  Our Interpretation is that saturation is less prominent if there are more availability variables: with a fully connected network, n-GateON has $N$ availability variables per layer, whereas p-GateON has  $N^2$.

We report the forgetting rate \textit{FR} in  Table \ref{tab:transferMNIST} (right column). Since all the values are positive it means that the model operates in the forgetting regime on this task and not in the backward transfer regime. Forgetting is less pronounced for $\epsilon=0$ and p-GateON which is consistent with the idea that saturation and forgetting exhibit antagonistic tendencies in this model.

\paragraph{Transfer learning on rotated and shuffled MNIST.}
We now ask whether our model is capable of inducing transfer learning. We first study forward transfer using the accuracy deviation $\Delta A_{cc}$  which compares continuous and isolated accuracy: a positive value indicated that the network benefited from previous tasks.
Our results are summarized in Table \ref{tab:transferMNIST} (left). All table entries showing a positive accuracy deviation (evidence for forward transfer learning) are shown in green.

We find that all GateON variants can achieve forward transfer learning. 
Interestingly, forward transfer occurs in the \textit{Rotated} and \textit{Shuffled} MNIST but not on the \textit{Permuted} MNIST.
It is on \textit{Rotated} MNIST with n-GateOn and $\epsilon=1$ that we achieved the highest forward transfer learning.
We interpret that forward transfer is easier when some structure is shared across tasks (e.g., visual features can be re-used in \textit{Rotated}, but not in \textit{Permuted} MNIST).

\begin{figure}[t!]
    \centering
    \includegraphics[width=1\linewidth]{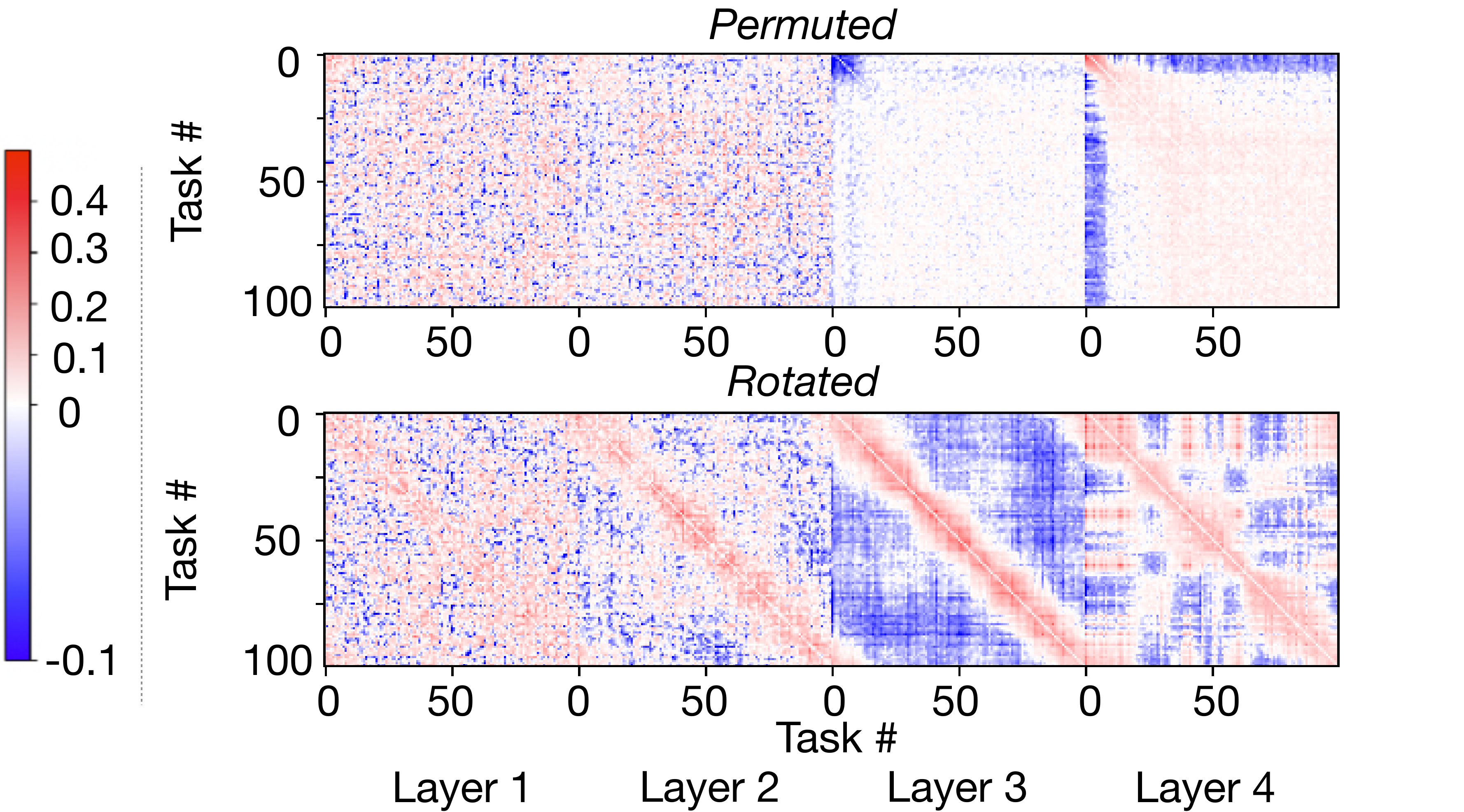}
    \caption[$\,$ Neuron sharing in GateON]{{\bf{Emergence of shared network structure through transfer learning}}. This figure showcases the Pearson correlations of context weights across all layers of  n-GateON$_{CNN}$ 0 after, trained on 100 tasks for  \textit{Permuted} (top) and \textit{Rotated} (bottom). We use n-GateON$_{CNN}$  to show the effect on both fully connected and convolutional layers. As expected for \textit{Permuted} after $15$ tasks we observe a positive correlation of the context weights in the output layers suggesting that it is then shared across tasks, in contrast, there is no structure in early convolutional layers due to the randomness of the permutation. 
    For \textit{Rotated} we re-ordered after training the tasks by increasing angles. The context weights are correlated for neighboring tasks to facilitate generalization. The correlation across neighboring tasks is most striking at intermediate layers.}
    \label{fig:correlations}
\end{figure}

To understand how transfer learning occurs we attempt to reverse engineer the trained network in Fig. \ref{fig:correlations}.
We display the Pearson correlation between gating weights matrices across the $100$ tasks in each of the four layers of the network. 
In \textit{Permuted} MNIST, after $20$ tasks, we see a high correlation in the output layer, suggesting that the final layer is shared across tasks (last panel of the first row in Fig. \ref{fig:correlations}), but no systematic correlation in earlier layers.
In the second row of Fig \ref{fig:correlations}, we show the same analysis for \textit{Rotated} MNIST. The tasks are re-ordered to ensure that neighboring task numbers correspond to similar rotation angles. In this case, we see strong correlations along the diagonal across
 all layers suggesting that the features are shared across tasks with neighbouring angles. 

\begin{table}[t!]
\centering
  \begin{tabular}{|l||l|l|l|}
    \hline
    & \textit{Permuted}& \textit{Rotated} & \textit{Shuffled}  \\
    \hline
    Vanilla gradient descent  & 26.3 & 69.7 & 13.4 \\
    %\hline
    %Active dendrites$^{\mathcal{R}}$\cite{iyer2022avoiding} & 91.2 & - & - \\
    \hline
    Obstruction only& 18.7 & 48.3  & 11.1  \\
    \hline
    Gating only & 55.1 & 71.6  & 62.1  \\
    \hline
    \hline
    n-GateON 0& 73.6 & 97.6  & {\bf{97.7}}  \\
    \hline
    n-GateON 0.5& 92.9 & 97.6 & 97.3   \\
    \hline
    n-GateON 1& 95.7 & 97.7 & 96.9 \\
    \hline
    \hline
    p-GateON 0& 96.3 & {\bf{97.9}}  & {\bf 97.7} \\
    \hline
    p-GateON 0.5& {\bf{96.5}} & 97.8  & 97.5  \\
    \hline
  \end{tabular}
\vspace{0.2cm}
 \caption[Results MNIST 100]{{\bf Ablation study for GateON on large task families.} Continual accuracy $A_{cc}^{cont}$  results (in percent) on \textit{Permuted}, \textit{Rotated}, and \textit{Shuffled} MNIST for 100 tasks.
 This harder setting with many tasks was only considered in the Active Dendrites paper \cite{iyer2022avoiding}, where they report $91.2$ continual accuracy on the permutated task. 
 Results for GateON with CNNs can be found in Supplementary  \ref{supp:task-inferredd} Table. %The values in bold indicate the best performance.
  %For $A_{cc}^{cont}$, green shows all the values that are within 1\% of the best performance.
  %For $\Delta A_{cc}$ and \textit{FR} the green values show forward (positive) and backward (negative) transfer, respectively.
  %On MNIST we do not observe backward transfer (no green).
  }
 \label{tab:ablation}
\end{table}

\begin{table}[t!]
\centering
  \begin{tabular}{|l| |l|l|l| |l|l|l|}
    \hline
     & \multicolumn{3}{c||}{{\bf Forward transfer}  $\Delta A_{cc}$} & \multicolumn{3}{c|}{{\bf Backward transfer} \textit{FR} }\\
    & \textit{Permuted}& \textit{Rotated} & \textit{Shuffled} &  \textit{Permuted}& \textit{Rotated} & \textit{Shuffled} \\
    \hline
    n-GateON 0 & -5.61 & \color{OliveGreen}0.59  & \color{OliveGreen}0.26  & 14.82 & 0.35 & {\bf{0.12}}  \\
    \hline
    n-GateON 0.5 & -1.57  & \color{OliveGreen}0.66 & \color{OliveGreen}0.39  & 2.06  & 0.33 & 0.62  \\
    \hline
    n-GateON 1 & -0.48 & {\bf \color{OliveGreen}0.7}  & \color{OliveGreen}0.4 &  1.12   & 0.24  & 1.09 \\
    \hline
    \hline
    p-GateON 0  & -0.6 & \color{OliveGreen}0.64 & {\bf \color{OliveGreen}0.24} & {\bf{0.48}}  & {\bf{0.14}} & 0.13  \\
    \hline
    p-GateON 0.5 & {\bf -0.05} & \color{OliveGreen}0.55 & \color{OliveGreen}0.29  & 0.84   & 0.16 & 0.37 \\
    \hline
  \end{tabular}
    \vspace{0.2cm}
  \caption[Results MNIST 100]{ {\bf Transfer learning on 100 MNIST tasks}. On the same simulation as in Table \ref{tab:ablation} we report the transfer learning capabilities of GateON. We measure forward and backward transfer learning with $\Delta A_{cc}$ and \textit{FR}, respectively. Green values show evidence of forward (positive) or backward (negative) transfer. On MNIST we do not observe backward transfer (no green).}
 \label{tab:transferMNIST}
\end{table}

\subsection{Continual learning in natural language processing}
Continual learning is  particularly relevant for language models for two reasons:
On a metaphoric level,  a convincing model of lifelong learning should explain the mechanism that enables learning new foreign languages without forgetting the one previously acquired.
On a more pragmatic level, a typical use case for CL in machine learning would be to update the knowledge of a pre-trained language model year after year as more data accumulates. In the present work we describe how to apply the GateON principle to language models of intermediate sizes and report the performance on existing CL benchmarks for NLP: Aspect Sentiment Classification (ASC) \cite{xu2019bert}, Document Sentiment Classification (DSC) \cite{ke2020continual}, and Text Classification using 20News data \cite{lang1995newsweeder} (20News). We summarize the three CL benchmarks below:

{\em{The ASC CL problem}} involves sentiment analysis on product reviews, with the network required to classify each aspect of the product, such as the picture or sound, as positive, negative, or neutral. The CL problem includes 19 products, with each product considered a separate task. The aspect term and sentence are concatenated using the [SEP] token, and the [CLS] token is used to predict the opinion polarity. For example, the network receives  “[CLS] torchlight [SEP] this is a good torchlight” and should output 'positive' for the aspect torchlight. Some reviews can have multiple aspects.

{\em{The DSC CL problem}} requires sentiment analysis on the full customer review, where the network has to classify the overall tone as positive or negative. The CL problem comprises 10 products (tasks), and the input comprises the token [CLS] followed by the text review.  The first output token is used as the sentiment estimate.

{\em{The split 20News CL problem}} is a topic classification task on the 20News dataset. The original task has 20 topics, which were randomly split into ten tasks, each containing two topics. The input consists of the token [CLS] followed by text.  The first output token is used as the topic estimate.

All these are CL-problems, meaning that their tasks are shown sequentially without recall. ASC and DSC are similar to \textit{Permuted} and \textit{Rotated} in the sense that each task has the same classification objectives but with different datasets, while \textit{Split} 20News changes the labels similarly as for \textit{Split} MNIST. We use both $A_{cc}^{cont}$ and Macro-F1 (MF1) score for comparison.  MF1 is the F1 score averaged over the different tasks and is more relevant than accuracy on biased datasets such as $ASC$.
To solve these CL problems we use one of the most used large language models, pre-trained BERT \cite{devlin-etal-2019-bert} to which we applied the GateON method on each module, attention, intermediate, output, for all 12 layers as well as pooling and embedding. 

\begin{table}[t!]
\centering
  \begin{tabular}{|l||l|l|l|l|l|l|}
    \hline
    \multirow{2}{*}{Models} &
      \multicolumn{2}{c|}{ASC} &
      \multicolumn{2}{c|}{DSC} &
      \multicolumn{2}{c|}{20News} \\
    & $A_{cc}^{cont}$ & MF1 & $A_{cc}^{cont}$ & MF1 & $A_{cc}^{cont}$ & MF1 \\
    \hline
    \makecell{BERT MTL \\ (max baseline)} & 91.9 & 88.1 & 89.8 & 89.3 & 96.8 & 96.8  \\
    \hline
    \hline
    BERT naive & 49.6 &43.1 & 73.1 & 71.8 & 52.5 & 39.2 \\
    \hline
    HAT\cite{zenke2017continual} & 86.7 & 78.2 &  87.3 &  86.1 &  93.5  & 92.9 \\
    \hline
    LAMOL\cite{sun2020lamol} &  88.9 &  80.6 &  {\bf{92.1}} & \bf{{91.7}} & 66.1 & 45.7 \\
    \hline
    B-CL\cite{ke-etal-2021-adapting} & 88.3 & 81.4 & 79.8 & 76.5 & 95.1 & 95.0 \\
    \hline
    CTR\cite{ke2021achieving} &  {\bf{89.5}}& \secbest{83.6} &  \secbest{89.3}  & 88.7 & 95.2 &95.2 \\
    \hline
    n-GateON 0 ({\bf ours}) & \secbest{89.3} & {\bf{84.1}}  & \secbest{89.3} & \secbest{88.8} &  \secbest{95.8}  &  \secbest{95.9}  \\
    \hline
    p-GateON 0 ({\bf ours}) & 83.1 & 76.7  & 87.1 &  85.3 & {\bf{95.9}}& {\bf{96.0}} \\
    \hline
  \end{tabular}
    \vspace{0.2cm}
\caption[Results for large language model]{{\bf Continual learning performance with language models.} For reference, we show the multi-task learning (MTL) baseline, all tasks are trained simultaneously instead of sequentially. We compare our results with the best models that we know of. To highlight that n-GateON is consistently within the top-2 algorithms across all tasks when compared to the models reviewed in \cite{ke2021achieving}, we highlighted with bold font the best performance in black and the second best in gray.}
 \label{tab:BERTresults}
\end{table}

\paragraph{GateON performance for fine-tuning language models.}
Remarkably n-GateON emerges as a versatile method that achieves competitive performance on all language-based CL problems (Table \ref{tab:BERTresults}).
When compared to all models for which the performance on this problem have been benchmarked by Ke et al. \cite{ke2021achieving}, we see that n-GateON is consistently the best or second best algorithm on all metrics.
This is remarkable because we used here only the two standard GateON Principles which are conceptually simple whereas  the other machine learning algorithms have often be designed specifically for CL on these tasks. For instance, both LAMOL and CTR which are the two most competitive alternatives to GateON are designed for BERT models and are not generalizable to other architectures.
The benchmark that is hardest for GateON appears to be DSC. There n-GateON is the second best algorithm behind LAMOL, but LAMOL achieves only a low performance on 20News suggesting that it is less versatile than GateON.
Interestingly, p-GateON achieves a relatively low performance on ASC and DSC in comparison with n-GateON. We believe that the parameter view may require more data to fine-tune robustly all relevant parameters. 

\paragraph{Forward and backward transfer in NLP}
As done previously with MNIST we report the accuracy deviation $\Delta A^t_{cc}$ and the forgetting rate \textit{FR} to study forward and backward transfer across tasks, respectively. 
In contrast to MNIST where all the tasks are designed with the same number of samples, the NLP benchmarks exhibit a wide range of task sizes, with some tasks characterized by limited and biased datasets.
In this sense, we expect that transfer across tasks is crucial for the tasks with few data points.

We report the $\Delta A^t_{cc}$ to study forward transfer in Table\ref{tab:forwardlll}. Both p- and n-GateON show positive accuracy deviations on all tasks emphasizing the method's capacity for forward transfer. It is most striking in ASC and DSC where the tasks are related to  specific "products", and a single product-specific dataset is sometimes small or very biased. The dataset Split 20news has consistent dataset sizes and highly independent tasks, which contribute to a more limited impact of transfer learning.

To study backward transfer in these benchmarks, we  report the values of forgetting rate \textit{FR} in Table \ref{tab:backwardlll}. While we were unable to see evidence for backward transfer on MNIST tasks previously, now we see that n-GateON has negative forgetting rates on $ASC$ and $DSC$, suggesting that the network improves its performance on previous tasks through backward transfer - without replaying the data from the previous task. Interestingly, n-GateON appears consistently better in terms of backward transfer than p-GateON, which highlights the importance of the neuro-centric view. 
%and since p-GateON maintains positive forgetting rate on all tasks we also conclude that it is consistently subject to catastrophic forgetting in this case.  

\begin{table}[t!]
\centering
\begin{tabular}{|l|ll|ll|ll|}
\hline
                        & \multicolumn{2}{c|}{ASC} & \multicolumn{2}{c|}{DSC} & \multicolumn{2}{c|}{20News} \\
                Models  & \multicolumn{1}{l|}{$\Delta{A{cc}}$} & \multicolumn{1}{l|}{$\Delta{MF1}$} & 
                              \multicolumn{1}{l|}{$\Delta A_{cc}$} & \multicolumn{1}{l|}{$\Delta{MF1}$} &
                               \multicolumn{1}{l|}{$\Delta A_{cc}$} & \multicolumn{1}{l|}{$\Delta{MF1}$}                
                             \\ \hline \hline
\multicolumn{1}{|l|}{n-GateON} & \multicolumn{1}{l|}{ \color{OliveGreen}{\bf 3.52}} & \multicolumn{1}{l|}{\color{OliveGreen}{\bf 6.28} }  & \multicolumn{1}{l|}{\color{OliveGreen}{2.34}}    & \multicolumn{1}{l|}{\color{OliveGreen}{2.99}}   & \multicolumn{1}{l|}{\color{OliveGreen}{\bf 0.16}}    & \multicolumn{1}{l|}{\color{OliveGreen}{\bf 0.17}}    \\ \hline
\multicolumn{1}{|l|}{p-GateON} & \multicolumn{1}{l|}{\color{OliveGreen}{2.24}}    & \multicolumn{1}{l|}{\color{OliveGreen}{1.99}} & \multicolumn{1}{l|}{\color{OliveGreen}{\bf 2.78}}    & \multicolumn{1}{l|}{\color{OliveGreen}{\bf 3.17}}  & \multicolumn{1}{l|}{\color{OliveGreen}{0.12}}    & \multicolumn{1}{l|}{\color{OliveGreen}{0.13}}  \\ \hline
\end{tabular}
\vspace{0.2cm}
\caption[Forward transfer in language models]{{\bf{Forward transfer in large language model}}. The deviation accuracy ($\Delta A_{cc}$) and MF1 deviation ($\Delta MF1$) are the performance improvement against the baseline where tasks are trained in isolation. All tasks show systematic positive deviation implying a forward transfer of knowledge for both n- and p-GateON. With an MF1 deviation up to 6.3\% for ASC. ASC comprises many small and biased datasets that seem to profit from this knowledge transfer.}
\label{tab:forwardlll}

\end{table}
\begin{table}[t!]
\centering
\begin{tabular}{|l|ll|ll|ll|}
\hline
        & \multicolumn{2}{c|}{ASC} & \multicolumn{2}{c|}{DSC} & \multicolumn{2}{c|}{20News} \\
        Models & \multicolumn{1}{l|}{$FR_{A_{cc}}$} & \multicolumn{1}{l|}{$FR_{MF1}$} & 
                              \multicolumn{1}{l|}{$FR_{A_{cc}}$} & \multicolumn{1}{l|}{$FR_{MF1}$} &
                               \multicolumn{1}{l|}{$FR_{A_{cc}}$} & \multicolumn{1}{l|}{$FR_{MF1}$}                
                             \\ \hline \hline
\multicolumn{1}{|l|}{n-GateON} & \multicolumn{1}{l|}{\color{OliveGreen}{\bf -0.21}}    &  \color{OliveGreen}{\bf -0.09}  & \multicolumn{1}{l|}{ \color{OliveGreen}{\bf -0.02}}    &  {\bf 0.03}   & \multicolumn{1}{l|}{\bf 0.08}    &  {\bf 0.08}   \\ \hline
\multicolumn{1}{|l|}{p-GateON}  & \multicolumn{1}{l|}{2.33}    &   1.16 & \multicolumn{1}{l|}{1.56}    &  2.32   & \multicolumn{1}{l|}{0.12}    &   0.12  \\ \hline
\end{tabular}
\vspace{0.2cm}
\caption[Backward transfer in language models]{{\bf{Backward transfer in large language model}}. The forgetting rate (FR) is a measure of the difference in accuracy between a model's performance immediately after training on a particular task and its performance after training on all subsequent tasks. For ASC FR is negative, showing that on average previous tasks learn more than they forget, showing the backward transfer capacity of n-GateON. In contrast, the p-GateON model has generally high FR values, which may explain its poor performance.}
\label{tab:backwardlll}
\end{table}

\section{Discussion}

\paragraph{Overview of the GateON theory.} In summary, we have described a model of plasticity relying on two principles: (1) Gated context selectivity by multiplicative modulation of the neuronal gain function; and (2) modulable freezing of weight updates if the parameters (or neurons) are relevant for previous tasks, yet with the possibility of unfreezing later one. This theory is implemented in two forms: a parametro-centric and a neuro-centric view. While the former provides a normative theory that links to previous successful models, we design the neuro-centric view as a biologically plausible model of continual learning in the brain.

\paragraph{Experimental prediction and implications for neuroscience}
The neuro-centric implementation of GateON provides a concrete and testable circuit-level hypothesis. Unlike previous theories of CL which are neuro-centric, our theory merges Principles 1 and 2 of CL into a unique and simpler hypothetical mechanism:  CL in the cortex is supported by the task selectivity of cortical neurons (Principle 1) and we predict the existence of neural availability variables that track the local activity and control (un-)freezing of neuronal plasticity when a neuron is selectively active on task $k$ (Principle 2). 
Our notion of availability is closely linked to the concept of metaplasticity \cite{Abraham08,el2012stable} but also to synaptic consolidation for which several models exist \cite{Fusi2005-fk,Zenke2014,Ziegler15,Benna2016-cf}.
The multiplicative gain function could be implemented by dendritic dynamics \cite{wybo2023nmda}.
%Despite the conceptual simplicity, this model is demonstrated to be  functionally useful and achieves state-of-the-art performance on machine learning benchmarks across various CL benchmarks.

%In our model, the availability for synaptic changes depends on the relevance 
%during earlier stimulations.
%Our biologically plausible relevance detection makes a clear 

A first prediction, based on the simplified relevance model of Eq. \eqref{eq-mu8}, is the following. Let us monitor the spiking activity of a neuron during an extended period of time. To turn spikes into a firing rate will require a low-pass filter which we estimate to be in the range of one second. Since we work with a normalized relevance, we focus on a neuron with a firing rate larger than the average firing rate of other neurons in the same area.  The  firing rate of the neuron should be squared to get the relvance, transformed into normalized relevance, and then inserted into Eq.  (\ref{eq:availabilty}) of the availability dynamics.  Our model predicts that freezing is visible as metaplasticity, i.e., a decrease in the amount of plasticity under a standard plasticity-induction protocol.  

Moreover, unfreezing is a second prediction of the model.
On a qualitative level, our
 model predicts that there is a threshold parameter $\epsilon>0$ such that if the firing rate is, for a long time, lower than the average firing rate of other neurons in the same area, then metaplasticity shows as unfreezing of plasticity. In other words, neuronal synapses can become plastic again.
At the moment the above statements are speculations, but potentially worth an experimental test.

%Overall, the GateON method has great potential for the field of continual learning, particularly in areas such as bio-inspired robotics\cite{LESORT202052}, where the ability to generalize from just a few real-world examples is crucial; or NLP where new datasets arise that might not be available at training time. As an online method, unlike replay-based methods, GateON does not require memory of previous datasets and avoids catastrophic remembering\cite{323080}. 

{\bf Comparison with other computational models of CL.} Four aspects are important to highlight similarities and differences to existing models.

(i) {\em Relevance estimation} is used in several other models, either in combination with gradient obstruction (e.g., HAT\cite{serra2018overcoming},
and RMN\cite{kaushik2021understanding}) or in combination with regularization to prevent large parameter drifts (EWC\cite{kirkpatrick2017overcoming}, SI\cite{zenke2017continual}). HAT uses an attention mask for relevance, which may overestimate the true importance of a parameter, while RMN uses an estimation of relevance learned during training. By contrast, we compute online an analytical estimate of the impact of each parameter on the loss of the {\em current} task which defines the instantaneous relevance of a parameter. 
 
 (ii) {\em Dynamic availability enabling freezing and unfreezing.} In contrast to existing availability control methods which freeze parameters of neurons of high relevance \cite{serra2018overcoming,kaushik2021understanding},
 GateON integrates the relevance gradually into a set of dynamic availability variables which determine whether, and how much,  parameters are allowed to change. A key advantage of our design over HAT or RMN \cite{serra2018overcoming,kaushik2021understanding}  is that it enables the introduction of a hyperparameter $\epsilon$ to control the unfreezing of parameters that have not been relevant for a long time while hedging against catastrophic forgetting. The dynamic availability variables could very well be related to metaplasticity, as explained above. 

 (iii) {\em Gating} has become a mainstream ingredient for efficient CL \cite{kirkpatrick2017overcoming,serra2018overcoming,masse2018alleviating,iyer2022avoiding,flesch2023modelling}. For instance, in contrast to HAT\cite{serra2018overcoming} which employs a binary attention mechanism for gating and a multi-head output, our approach gates every unit continuously and does not require a multi-head output. Our gating approach shares similarities with the learnable context masking in the Active Dendrites model \cite{iyer2022avoiding}, but in contrast to their approach, we do not enforce a strict number of active neurons per context nor do we need to estimate a task embedding.

 (iv) {\em Automatic context detection}. Whereas standard CL models give explicit information on task numbers \cite{kaushik2021understanding}, we introduce a simple context detector that enables unsupervised inference of tasks. 
 While the fields of machine learning and signal processing have incorporated context switches via out-of-distribution detection \cite{Adams07,Fearnhead07,Nassar10,nalisnick2019detecting,Liakoni21}, we argue that there exist simple context detectors like ours, which are computationally cheap since it focuses on deviations in the loss which is evaluated anyway for each sample. Other context detectors were studied in \cite{caccia2020online,heald2021contextual}.

{\bf Conclusion.} The paradigm of CL is different from standard statistical learning since data arrives as an input stream with a non-stationary data distribution, potentially characterized by sharp switches between tasks or contexts. We have shown that the neuro-centric view of gating feedforward processing in combination with metaplasticity of learning rules has foundations in the neurosciences and leads to state-of-the art performance on established, and novel, machine learning problems. Due to its structural similarity to life-long learning in biology, CL
will also in the future continue to trigger a fruitful crosstalk between neuroscience and machine learning.

%\paragraph{} In conclusion, the GateON method is a powerful continual learning approach that can be applied to many types of neural networks by incorporating a context-gating layer and gradient flow obstruction. Unlike previous models \cite{zenke2017continual,kaushik2021understanding,serra2018overcoming}, we showed that our approach is general enough to be used either with relevance detection on the level of neurons or individual  parameters. Of minimal algorithmic complexity, the method can prevent catastrophic forgetting by analytically computing the relevance of each neuron (or parameter) online and adjusting their learning rate.
%Here the notion of online is meant in opposition to replay methods, but can include mini-batches of, say, 1 to 1000 data points.
%On MNIST CL tasks p-GateON performed better than n-GateON, because of the large amount of available training data. On the other hand, n-GateON performed exceptionally well on NLP continual learning tasks using pre-trained transformers. It has also shown positive forward transfer capacity for all tasks. Our experiments on the ASC dataset revealed a negative forgetting rate, indicating that the method possesses backward transfer capabilities for incomplete datasets. 

\section{Methods}

\subsection{Task inference}
\label{matmet:taskinference}
In this section, we present an unsupervised algorithm to identify change points and infer the current task.  % While our approach employs a single context for the entire network, it is adaptable for application to individual network layers. 

Before we turn to change-point detection, we start by defining the concept of network confidence in the current context. Denoted as $\text{confidence}_k$, the confidence quantifies the network's level of expertise regarding context $k$. This confidence is updated as follows:
\begin{equation}
\Delta \text{confidence}_k =  \eta_C \, [C_k - \text{confidence}_k ]
\end{equation}
where $\eta_C$ is a hyperparameter and $C_k\in \{0,1\}$ is the value of the context layer under one-hot encoding.  In other words, the confidence for the context $k$ increases (over a time scale of roughly $1/\eta_C$ if the said context is active during the backward pass.

For each data point (or minibatch of data), we evaluate the loss  $\mathcal{L}^n)$ (where the index $n$ denotes an unknown current task label) and its running average
\begin{equation}
\Delta \bar{\mathcal{L}} = \eta_L(-\bar{\mathcal{L}}+\mathcal{L}^t)
\end{equation}
where $\eta_L$ is a parameter. The running average smoothes the loss over a time scale of roughly $1/\eta_L$.

\begin{algorithm}[t!]
\caption{Change-point detection algorithm for GateON}\label{matmet:alg:euclid}
\begin{algorithmic}[1]
\Procedure{CPD}{}

\State $ \text{Change-point} \gets  false$
\State $ \Theta, \eta_C, \eta_L,m \gets \text{set hyper parameters}$ \Comment{ $m>1/\eta_L$ controls window of min-search}
\State active-context $\gets 0$
\State Confidence(active-context) $\gets 0$
\State $N_{contexts} \gets $ 1
\For{data in datasets}
\State $x^{0} \gets data$
\State $\mathcal{L}^t \gets forward(x^{0}, \text{active-context})$\Comment{forward pass network}
\State $\bar{\mathcal{L}} = \bar{\mathcal{L}} + \eta_L(-\bar{\mathcal{L}}+\mathcal{L}^t)$
\If{$\min\limits_{t-m \leq t' \leq t}\mathcal{L}^{t'}> \Theta \cdot \bar{\mathcal{L}}  \text{ and Confidence(active-context) } >0.9$}
\State reactivated $\gets$ false \Comment{Trying to reactivate previous context}
   \For{test\_context in $N_{contexts}$}
        \State $\mathcal{L}^t \gets forward(x^{0}, \text{test\_context})$
        \If {$\mathcal{L}^t< \Theta \cdot \bar{\mathcal{L}}$}
        \State active-context $\gets$ test\_context
        \State reactivated $\gets$ True
        \State Break
        \EndIf
    \EndFor
\EndIf
\If {not reactivated}\Comment{if no previous context suitable}
\State active-context $\gets$$ N_{contexts}$
\State $N_{contexts}$ $\gets$ $N_{contexts}$ + 1
\State  Confidence(active-context)$ \gets 0$
\EndIf
\State Change-point $\gets$ false
\State Confidence(active-context) $\gets $Confidence(active-context) $(1- \eta_C) + \eta_C$
\State Continue the training steps...
\EndFor
\EndProcedure
\end{algorithmic}
\end{algorithm}

After these preparations,  we are ready to describe the change-point detection algorithm in ALG \ref{matmet:alg:euclid}. The basic idea is that we detect change points by comparing the current loss to the average one in the present context.
To increase stability, the 'current' loss is the minimum loss over the last $m$ steps. Specifically, 
\begin{itemize}
 
\item In line 11, we compare the minimum loss over the last $m$ steps to a low-pass filtered loss. A change point is detected if the error exceeds a predefined threshold $\Theta$ and the network exhibits sufficient confidence; otherwise, the network remains in its current context. We consider the minimum of the last $m$ steps to be sure that the current loss value is not an outlier. The larger $m$ the longer to detect a change-point but the more sure we get.

\item Upon detecting a change point, lines 12-18 assess if one of the existing  contexts $N_{contexts}$ aligns well with the present observation. If such an alignment is found, the previous context is reactivated; if not, a new context is generated and the counter $N_{contexts}$ is increased by one.

\item This procedure is carried out at every time step (where time step means one data point or one minibatch).
\end{itemize}

% While we have reported 100\% accuracy for the task presented in this paper, for the sake of computation time, we fed the task directly. However, the algorithm has been implemented in our code and contains two modes: task-fed, and context-inferred network. 

\begin{figure}[t!]
\centering
\includegraphics[width=.8\linewidth]{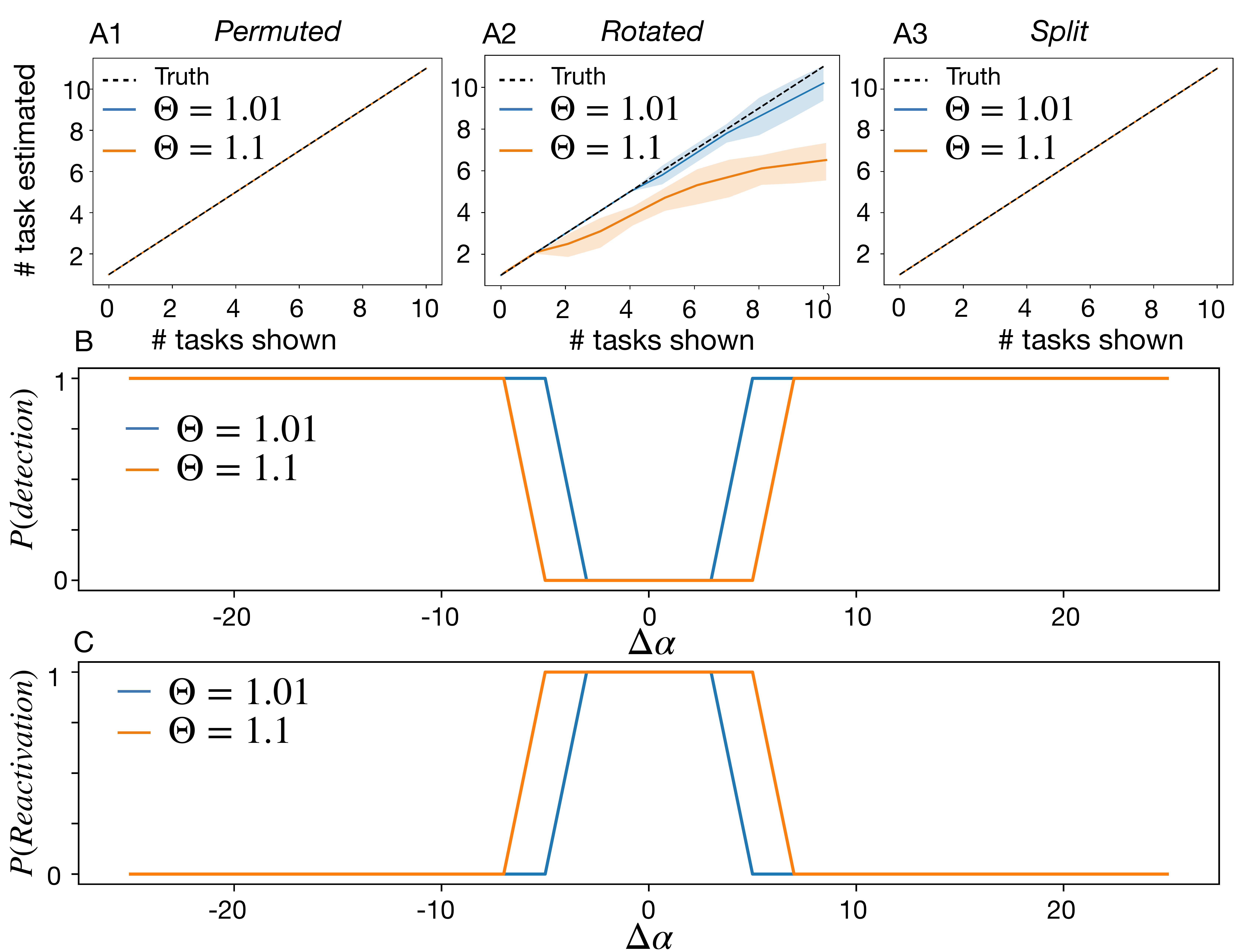}
\caption{{\bf{Context Detection in MNIST}}. {\bf{A1-3}} We executed 10 continual learning (CL) tasks and tracked the number of tasks identified by GateON compared to the number of tasks presented. The results are plotted for two $\Theta$ values, where $\Theta$ modulates task detection stringency as explained in Algorithm \eqref{matmet:alg:euclid}. The dashed black line represents perfect task estimation accuracy. Results are averaged over 10 trials, with standard deviation indicated by the shaded region. {\bf{A1}} and {\bf{A3}} exhibit parity between the number of tasks presented and detected for both $\Theta$ parameters. In contrast, {\bf{A2}} reveals that for \Rm, GateON underestimates the actual task count, with larger $\Theta$ values leading to fewer tasks detected. Specifically, tasks with closely spaced angles may be misconstrued as a single task, leading to context collapse. {\bf{B}} To elucidate the reasons for the incomplete task detection in {\bf{A3}}, we executed two consecutive \Rm tasks with an angle difference $\Delta \alpha \in [-25,25]$, and charted the switch-point detection probability for GateON (averaged over 10 trials for each $\Delta \alpha$). The results indicate a binary behavior: if $\Delta \alpha$ exceeds a certain threshold, tasks are distinctly identified with a probability of 1; otherwise, tasks are amalgamated with equal certainty. This suggests a 100 percent switch-point detection rate if the angle difference is sufficiently large; otherwise, GateON perceives the tasks as identical. To confirm that similar tasks can reactivate prior ones, we ran in {\bf{C}} a sequence of three tasks such that angles 1 and 3 differ by $\Delta \alpha$ and angle 2 is always far from the other two to be detected as switch-point. {\bf{C}} demonstrates that the angle pairs leading to context collapse in {\bf{B}} will also trigger reactivation, while those that resulted in distinct tasks do not reactivate. In summary, our context detector appears proficient at identifying switch-points between significantly dissimilar tasks and can also reactivate or collapse tasks when they are sufficiently alike.}
\label{matmet:fig:taskinference}
\end{figure}

The use of inferred context can have benefits, as it allows us to set our tolerance for change points. For instance, in the case of \textit{Rotated}, the representational drift depends on the angle between task $k$ and $k'$, which means that we can choose a tolerance $\Theta$ at which we consider the tasks to be similar or dissimilar. In Fig. \ref{matmet:fig:taskinference} we study the properties of the task detection with MNIST CL tasks.

\subsection{Derivation of a bio-plausible relevance measurement}
\label{matmet:layer-wise-derivation}
For a biologically plausible relevance estimation, we consider n-GateON. However, computing the term  $\frac{\partial\mathcal{L}}{\partial x^l_i}$ might be impossible for biological neural networks since it requires to backpropagate detailed gradient information from the output layer to an arbitrary layer $l$. Below we provide a theoretical justification for the approximation $\mu_{x^l_i} = (x^l_i)^2$ which connects with the calculation of the causal of $x_i$ on deeper layers.

We propose a sequence of two approximation steps. The first one is a per-layer definition of causality. Instead of estimating the relevance of a neuron by its impact on the output layer,  we compute how much removing neuron $i$ in layer $l$ impacts the activity $x_j^{l+1}$ of neurons $j$ in the {\em next} layer $l+1$ and define
the relevance
$\mu_{i}^l =\frac{1}{N}\sum_j(a^{l+1,-i}_j - a^{l+1}_j)^2$ \,
 where $a^{l+1}_j = \sum_i w_{ji} x_i^l$ is the activation of the next layer before the non-linearity and the gating, $-i$ denotes that we removed the neuron $i$ of layer $l$ in the forward path.
%We exploit that the activity of $x_j^{l+1}$ is a product of a gating factor and the feedforward activity and we do not need to use a Taylor 
Since $a_j^{l+1} $ is linear in ${x_i^l}$ we have:
\begin{equation}
\begin{split}
    \mu_{i}^l =& \frac{1}{N}\sum_j [\sum_{i'\ne i} w_{ji'}^{l+1} x_i^l - \sum_{i} w_{ji}^{l+1} x_i^l]^2\\
    =&\frac{1}{N}\sum_j [ w_{ji}^{l+1} x_i^l ]^2, \\
    =& \frac{\sum_j {(w_{ji}^{l+1}})^2}{N} \cdot {x_i^l}^2 
    \,\,\label{eq:causallocal}
\end{split}
\end{equation}
 where the sum runs over all %active
 neurons in layer $l+1$.
 This layer-wise variation yields a formalization of the relevance that is free from the term $\frac{\partial L}{\partial x_i}$ which may require implausible network computations like Backprop. To simplify the computation further, we can observe that the term $\sum_j {w_{ji}^{l+1}}^2$ does not depend on the network activity and it is the norm of the weight vector leaving neuron $i$. If we further ignore the proportionality constant $\frac{\sum_j {(w_{ji}^{l+1}})^2}{N}$, equation \eqref{eq:causallocal} has the same effect as our most biologically plausible approximation of $\mu_i^l$:
 \begin{equation}\label{eq:bio-relevance}
     \mu_{i}^l = ({x^l_i})^2.
 \end{equation}
 For comparison with also derived the relevance computation $\mu_{i}^l =\frac{1}{N}\sum_j(x^{l+1,-i}_j - x^{l+1}_j)^2$ where the gating and the non-linearity of layer $l+1$ are taken into account. To be computable, we needed to make a first-order Taylor approximation to evaluate this term. It results in a more complex computation than Eq. \eqref{eq:bio-relevance} requiring a downward communication from layer $l+1$ to $l$. In simulations, we find a small benefit for this more complex approximation (orange line in Fig. \ref{matmet:fig:comparisonava}), but we did not explore this further.

\subsection{n-GateON for CNN}
\label{matmet:CNNGateON}
In the context of a convolutional neural network (CNN), it is necessary to define the gating and availability operations at the level of filters as opposed to individual neurons. Building upon the notation established in the previous section, we refer to the filters in the $l$-th layer of channel $c$ as $w_{a, b}^{c, l}$, where $a$ and $b$ correspond to the indices of the filter weights. By applying gating and availability operations to each filter in every channel $c$ of layer $l$, we can compute the corresponding gated activity as follows:
\begin{equation}
\begin{split}
x_{ij}^{l,c} &= g_c^{l}f(\sum_{a=0}^K\sum_{b=0}^Kw_{ab}^{c,l}x^{l-1,c}_{(i+a)(j+b)})\\
    \text{with }  g^{l}_c &= \sigma(v^{l}_{ck}),
\end{split}
\end{equation}

For p-GateON, the relevance mappings are exactly the same as defined for a fully connected network (one per parameter in the convolution kernel). With n-GateON we have one unit per channel per feature map location, which adds up to a large number of units per convolutional layer. It is, however,   necessary to define how freezing the parameter of a unit in one location freezes the shared parameter in other units of the same channel, across the convolutional layer.
For simplicity, and in agreement with the gating variable which acts similarly at all locations, we assign one relevance and availability variable per channel which are shared across the feature map. For instance using the bio-plausible approximation we have:
\begin{equation}
\begin{split}
    \mu_x^{l,c} &= \frac{1}{N_i}\frac{1}{N_j}\sum_i \sum_j (x_{ij}^{l,c} )^2
\end{split}
\label{eq:availabilty_conv}
\end{equation}
\subsection{Accuracy measures for Continual learning}
\label{matmet:measures}
We train the tasks $T_1, T_2, \dots, T_K$ sequentially. For any task $T_{k'}$  we can measure the accuracy on task $T_{k'}$ at time $t_k$, i.e., right after training on task $T_k$.
This test accuracy is denoted $A_{cc}^{k'}(t_k)$. The immediate test accuracy after training task $T_k$ being, $A_{cc}^{k}=A_{cc}^{k}(t_k)$.\\

After training on all tasks we measure the continual accuracy
\begin{equation}
    A_{cc}^{cont,k} = \frac{1}{K-k+1}\sum_{k'\geq k}^{K}
    A_{cc}^{k} (t_{k'}).
    \label{eq:accont}
\end{equation}
The continual accuracy is  the basis to define the forgetting rate FR$^k$.

To probe potential network saturation or forward transfer during training we measure the normalized difference  between the immediate test accuracy $A_{cc}^{t}$ of the GateON model and and isolated network. This yields the 
the relative immediate test accuracy ratio $\Delta A_{cc}^{k}$. We then average over all tasks 
\begin{equation}
    \Delta A_{cc} = \frac{1}{\bar{A}_{cc}^{\text{Isolated}},}\left(\frac{1}{K} \sum_{k}^{K} A_{cc}^{k}
    -\bar{A}_{cc}^{\text{Isolated}}\right),
    \label{eq:accontrel}
\end{equation}
with $\bar{A}_{cc}^{\text{Isolated}}$ the average accuracy over all tasks if networks are trained separately for each task.  A positive $\Delta A_{cc}^{k}$ shows that on average the immediate test accuracy is better with CL  (forward transfer) than without.
In the Results section, we present results in \%, i.e. multiplied by 100. 

\subsection{Task-locked accuracy}
\label{matmet:tasklocked}
While in continual accuracy and related measures we focus on the performance at times $t_k$, i.e., at the end of a task $T_k$, for the task-locked accuracy we study the evolution of accuracy  {\em during} one task.  For MNIST we train each task for 500 minibatch steps (around 9 epochs) and then switch to the next one. During training, we save the test accuracy of task $T_k$ as $A^{k}_{cc}(step)$ for $step= t-t_{k-1} \in \{1,500\}$. \\ 
To avoid showing redundant plots (up to 100 tasks trained) and highlight the average convergence speed and accuracy during training we compute the average tasked-locked accuracy, 
\begin{equation}
    A_{cc}(step) = \frac{1}{K-1} \sum^{K}_{k>1}A^{k}_{cc}(step).
\end{equation}
$A_{cc}(step)$ highlights the average speed of training of tasks $T_2, T_3, ... $  after a task switch. 
%A fast re-training compared to $A^{1}_{cc}(step)$ shows the capacity of the network to generalize previous knowledge to new tasks. Similar convergence to the $A^{1}_{cc}(step)$ shows that the method does not harm the final accuracy either. In some cases, the final accuracy can even be better as the generalization might also benefit.

\subsection{Spefications of the MNIST CL benchmarks}
\label{matmet:MNIST}

\textit{Permuted} is a test for random input changes because the input pixels are randomly permuted, thereby rendering it impossible for a standard model to generalize across tasks. On the other hand, in \textit{Rotated}, if the rotation angle between two tasks is small, some visual features can be re-used for different rotation angles.
 
\textit{Split} is a test that is often used in CL papers and consists of grouping the dataset into five binary classification tasks corresponding to the detection of two random digits.  \textit{Shuffled} introduces random output changes (permutation of the 10 labels). This task is particularly challenging for classical networks (without multi-head) as it needs to override in the final layer the previous knowledge at each new task. In these tasks it is almost always impossible to generalize across tasks: for each new task inputs that represent a '3' must suddenly be mapped to another output unit coding for another digit. If the network does not have a task-specific output layer (also called network 'head' below) the weight of the deeper layer(s) has to be overridden to change the output convention.

\begin{figure}[t!]
    \centering
    \includegraphics[width=1\linewidth]{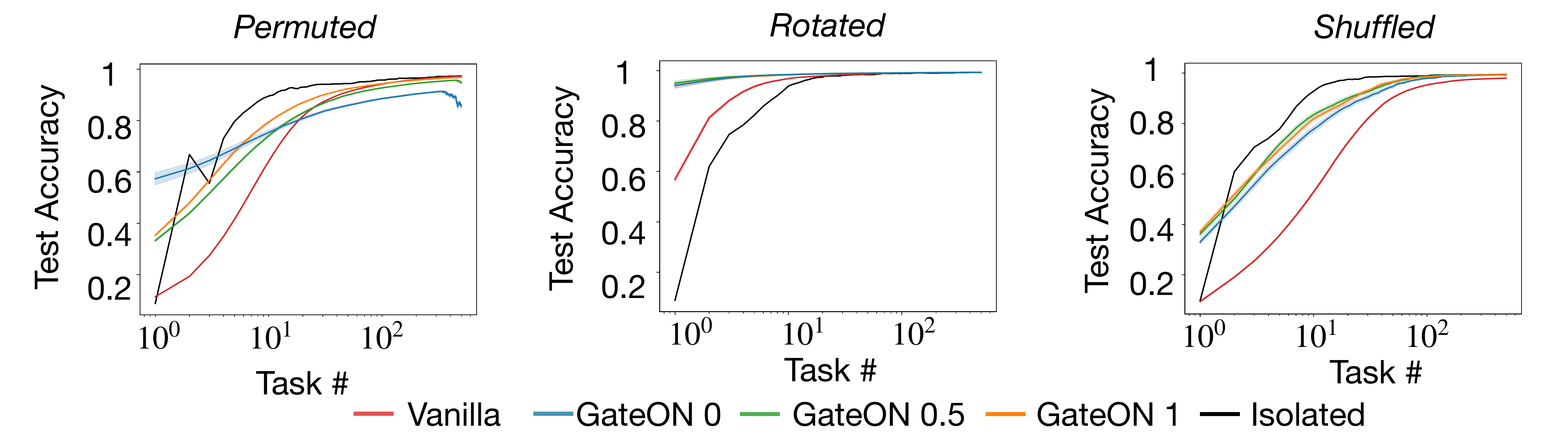}
    \caption[$\,$ Task-locked accuracy in GateON]{{\bf{Network generalization}}. The three figures show the task-locked accuracy ({\bf{Methods}} \ref{matmet:tasklocked}) over 500 batch presentations for all three MNIST tasks. The task-locked accuracy reveals the average re-training speed and final accuracy of the network. A network that generalizes well should learn faster and achieve higher accuracy if possible. Isolated indicates the average speed of training using a new network for each task. The number following GateON in the legend corresponds to the value of epsilon in Eq. \eqref{eq:availabilty} used for the GateON method.}
    \label{fig:mean_abs_acc}
\end{figure}

\subsection{Context correlation matrix}
\label{matmet:correlmat}
To be sure that sub-networks generated from the GateON method are not mutually exclusive we want to measure how the context weights that gate the activity of the neurons correlate between each task. To do so we binarise the context weight vector from context $i$ to all neurons in layer $l$
\begin{equation}
        w^{l,binarised}_{ik} =
    \begin{cases}
      1 & \text{if } w^{l}_{ik}>0\\
      0 & \text{otherwise.}
    \end{cases}
\end{equation}
We chose to binarise since we want to focus only on whether this neuron is used or not and hence if the neuron feature is re-used. Then we compute the Pearson correlation \cite{pearson1895vii}:
\begin{equation}
    C_{ij} = \frac{COV(\Vec{w}^{binarised}_i,\Vec{w}^{binarised}_j)}{\sigma_i\sigma_j}.
\end{equation}
Strong correlation means that similar neurons are used between the two tasks and hence the sub-networks are not separated. We expect this kind of behavior in correlated tasks like \textit{Rotated} with two close angles. In orthogonal tasks we would expect the neurons to be anti-correlated, i.e. the network actively chooses other neurons than the one previously used.

\subsection{Network architectures and training hyperparameters}
\label{matmet:architectures}

\begin{table}[!ht]
\begin{minipage}{0.45\textwidth}
    \centering
    \begin{tabular}{|c|c|c|c|c|}
    \hline
     & $m$ & $\Theta$ & $\eta_L$ &$\eta_C$\\
    \hline
    \Pm & 1 & 1.01 & 0.02 & 0.02 \\
     \hline
    \Rm  & 1 &  1.01 & 0.02 & 0.02\\
    \hline
    \textit{Shuffle}  &1 & 1.01 & 0.02 & 0.02\\
     \hline
    \end{tabular}
    \caption{GateON parameters for MNIST task detection}
    \label{tab:paramsMNIST}
\end{minipage}
\begin{minipage}{0.45\textwidth}
    \centering
    \begin{tabular}{|c|c|c|c|c|}
    \hline
     & $m$ & $\Theta$ & $\eta_L$ &$\eta_C$\\
    \hline
    ASC & 3 & 1.1 & 0.05 & 0.02 \\
     \hline
    DSC & 3 &  5 & 0.05 & 0.02\\
    \hline
    newsgroup & 3 & 5 & 0.05& 0.02\\
     \hline
    \end{tabular}
    \caption{GateON parameters for NLP task detection}
    \label{tab:paramsNLP}
\end{minipage}
\end{table}

\subsubsection*{10 MNIST Tasks}
For \Pm, the network is composed of one input layer of size 728, 2 hidden layers with $2000$ neurons, and an output layer of 10 neurons. We added two layers of conv with size 32 for \textit{Split} and 256,512 for \Rm. Each task was trained for 250 training steps with batch size 1000 (about 4 epochs).
% 2000 batch size
We use the Adam optimizer \cite{kingma2014adam} with a learning rate of $5e-3$, reset after each task switch. $\eta_A = 0.01$ for all layers. The gating activation function is a rectified hyperbolic tangent, ReLU for the neurons and softmax for the output. To make the learning faster we also add a batch norm at each layer. We use cross-entropy loss. We use a batch size of 1000.  Everything is implemented with Pytorch\cite{NEURIPS2019_9015}.
\subsubsection*{100 MNIST Tasks}
\paragraph{Fully-connected architecture and training}
The architecture is the same for all three CL problems with the same hyper-parameters using Pytorch \cite{NEURIPS2019_9015}. It is composed of one input layer of size 728, 4 hidden layers with $5000$ neurons, and an output layer of 10 neurons. The increase in the number of neurons (compared to the network for 10 tasks) has been necessary for solving 100 tasks. 
We use the Adam optimizer with a learning rate of $5e-3$, reset after each task switch. $\eta_A = 0.01$ for all layers.  The gating activation function is a rectified hyperbolic tangent, ReLU for the hidden neurons and softmax for the output. To make the learning faster we also add a batch norm at each layer, and we re-normalize the output layer activity at each layer. We use a cross-entropy loss. We use a batch size of 1000 and train each task for 50 epochs. 
% \begin{equation}
% \mathcal{L} = CE(o,t) + \lambda\sum_{l}\frac{1}{N}\sum_{i} g_i^{{l}},
%     \label{matmet:eq:loss}
% \end{equation}
% with $o$ the output and target $t$. The regularizer sparsifies the gating in the layers. $\lambda$ is 0.05 for all layers.

\paragraph{Convolutional architecture and training}
On MNIST with convolutional networks, the architecture is identical to \cite{NEURIPS2019_9015} for all three CL problems with the same hyper-parameters. It is composed of one input layer of size 728, 2 convolutional layers with 256 and 512 channels followed by one max pool of kernel 4, 2 hidden layers with $5000$ neurons, and the output layer of 10 neurons.
We use the Adam optimizer with a learning rate of $5e-3$, reset after each task switch. $\eta_A = 0.004$ for all conv layers, $\eta_A = 0.01$ all hidden layers. The gating activation function is a rectified hyperbolic tangent, ReLU for the neurons and softmax for the output. To make the learning faster we also add a batch norm at each hidden layer, and we re-normalize the output layer activity.
We use a cross-entropy loss. We use a batch size of 1000 and train each task for 50 epochs. 
\subsubsection*{CIFAR-100 Tasks}
\paragraph{Wide-resnet architecture and training}
We used Wide-ResNet as described in \cite{Zagoruyko2016} with a depth of 28 and a widening factor of 10. We changed the first convolution from 16 to 128 to avoid fixing too quickly the parameters. We use the SGD optimizer with a learning rate of $0.02$ and momentum of 0.9, with a cosine annealing scheduler. $\eta_A = 0.004$ for all conv layers, except for the first one where $\eta_A = 0.04$. The gating activation function is a rectified hyperbolic tangent, ReLU for the neurons and softmax for the output. To make the learning faster we also add a batch norm at each hidden layer, and we re-normalize the output layer activity.
We use a cross-entropy loss plus the mean absolute value of the context weights as a regularizer for the number of active channels.
We use a batch size of 128 and train each task for 50 epochs. 
\subsubsection*{Natural language processing tasks}
\paragraph{BERT network architecture}
We use the  BERT base model (uncased)  similarly as in Devlin et al.\cite{devlin-etal-2019-bert}. The same parameters are used for all tasks for the training. The parameters for context detection differ between tasks. The only adaption we did was to add the context layers. We also put the availability of the embedding layer and self-attention layers to 0. We do this because we did not have access to the pre-training to implement our relevance mapping and availability fixing. Since Attention and embedding are projections in small dimensions it has a very dense representation leading to most of the neurons being relevant. We expect that for these layers our model would fix most of the neurons during training.  Hence we set the availability of the embedding layer and each attention layer to 0. Note that they still have the gating that allows task-driven adaptation. 
The context parameters $w$ are initialized so that $\sigma(w) = 1$ for all neurons and contexts. It prevents gating to change too much the Bert representation before training. We use $\eta_A$ = 0.05 except if the first dataset has less than 50 batch updates in this case the first task uses $\eta_A = 0.1$.  We use a batch size of 32 and train each task for variable epochs depending on the dataset size as described in \cite{ke2021continual}.

\section*{Data and code availability statement}
The code is made publicly available at \url{https://github.com/martinbarry59/GATEONMNIST}.
All dataset used in this study are already publicly available.

\section*{Acknowledgement}
M.B. has conceptualized the research topic and pursued the investigation by himself (both simulations and analytical work). After the first draft was written, W.G. and G.B. helped to consolidate the analytical work and to write the final manuscript. 
This research was supported by the Swiss National Science Foundation with grant No. $200020\_207426$. The research of W.G. and G.B. was supported by a Sinergia grant with grant No. $CRSII5 198612$.

\bibliographystyle{unsrt}
\bibliography{main} 
\newpage
\appendix
%\addcontentsline{toc}{section}{Supplementary Material} % Add the appendix text to the document TOC
%\part{Supplementary Tables} % Start the appendix part
%\parttoc % Insert the appendix TOC

 %%% Uncomment this line and comment out the ``thebibliography'' section below to use the external .bib file (using bibtex) .

\newpage
\section{Supplementary tables}

\subsubsection*{Results GateON, task-fed} 

\begin{table}[!ht]
\centering
  \begin{tabular}{|l||l|l|l|l|l|l|l|l|l|}
    \hline
    \multirow{3}{*}{Models} &
      \multicolumn{3}{c}{\textit{Permuted}} &
      \multicolumn{3}{c}{\textit{Rotated}} &
      \multicolumn{3}{c|}{\textit{Shuffled}} \\
    & $A_{cc}^{cont}$& $\Delta A^t_{cc}$ & FR. & $A_{cc}^{cont}$& $\Delta A^t_{cc}$ & FR & $A_{cc}^{cont}$ & $\Delta A^t_{cc}$ & FR \\
    \hline
    Vanilla  & 26.29 & 0.0 & 71.57 &69.68 & 0.0 & 28.53 &13.41 & 0.0 & 77.44\\
    \hline
    \hline
    % Obstruct 0& 18.69 & 34.15  & 45.82 & 48.29 & 63.85  & 29.36 & 11.15 & 13.67  & 72.16 \\
    % \hline
    % Obstruct$_{CNN}$ & 16.16 & -51.17 & 76.96 &95.07 & -1.78 & 3.73 & 15.15 & 13.67  & 72.16 \\
    % \hline
    n-Gating & 55.07 &  {\bf{0.24}} & 41.96 & 71.63 &   0.94 & 25.33 & 62.07 & 0.6 & 35.11\\
    \hline
    \hline
    n-GateON 1&95.96 & 0.2 & 1.5 &97.46 & {\bf{1.18}} & 0.49 &96.38 & {\bf{1.05}} & 1.52 \\
    \hline
    n-GateON 0.5& 95.88 & -1.39 & 0.25  &97.6 & 1.08 & 0.19 &97.66 & 1.01 &  0.19 \\
    \hline
    n-GateON 0& 92.38 & -5.0 & 0.25 &97.39 & 0.82 & {\bf{0.05}} &{\bf{97.72}} & 0.93 & 0.06 \\
    \hline
    \hline
    p-GateON 0.5& 96.54 & -0.05 & 0.79 &{\bf{97.83}} & 0.67 & 0.14 &97.52 & 0.31 & 0.36  \\
    \hline
    p-GateON 0&{\bf{ 96.56}} & -0.62 & {\bf{0.23}} &97.8 & 0.64 & 0.06 &97.69 & 0.25 & {\bf{0.15}}  \\
    \hline
  \end{tabular}
  \caption[ $\,$ results MNIST 100, task fed]{Performance results on the three CL-tasks (\textit{Permuted}, \textit{Rotated}, and \textit{Shuffled}) for 100 tasks with task fed in the context layer.}
 \label{tab:suppaccMNIST-fed}
\end{table}

\label{supp:task-fed}
\begin{table}[!ht]
\centering
  \begin{tabular}{|l||l|l|l|l|l|l|l|l|l|}
    \hline
    \multirow{3}{*}{CNN Models} &
      \multicolumn{3}{c}{\textit{Permuted}} &
      \multicolumn{3}{c}{\textit{Rotated}} &
      \multicolumn{3}{c|}{\textit{Shuffled}} \\
    & $A_{cc}^{cont}$& $\Delta A^t_{cc}$ & FR. & $A_{cc}^{cont}$& $\Delta A^t_{cc}$ & FR & $A_{cc}^{cont}$ & $\Delta A^t_{cc}$ & FR \\
    \hline
    Vanilla  &41.32 &   & 55.91 &94.06 &   & 5.33 &40.15 &   & 58.22 \\
    \hline
    Gating & 50.03 & {\bf{-0.07}}& 47.15 &88.76 &   -0.04  & 10.1 & 48.63 & -0.14 & 50.65  \\
    \hline
    n-GateON 1  & 96.16 & -0.58 & 0.9 &98.83 & {\bf{0.43}} & 0.37 &98.5 & {\bf{-0.1}}2 & 0.78 \\
    \hline
    n-GateON 0.5  & {\bf{96.57}} & -0.72 & {\bf{0.37}} &98.62 & 0.38 & 0.36 & 98.97 & -0.15 & 0.28 \\
    \hline
    n-GateON 0 & 89.43 & -7.91 & 0.78 & {\bf{99.11}} & 0.25 & {\bf{0.06}} & 98.97 & -0.26 & {\bf{0.2}}  \\
    \hline
    \hline
    p-GateON 0.5 & 94.75 & -1.4 & 1.62 &99.1 & 0.11 & 0.21 &98.94 & -0.14 & 0.34   \\
    \hline
    p-GateON 0 & 95.02 & -2.34 & 0.48 &99.05 & 0.09 & 0.12 & {\bf{99.06}} & -0.14 & 0.22  \\
    \hline
  \end{tabular}
  \caption[ $\,$ results MNIST 100, task fed, with convolutions]{Performance results on the three CL-tasks (\textit{Permuted}, \textit{Rotated}, and \textit{Shuffled}) for 100 tasks with task-fed in the context module using CNN.}
 \label{tab:suppaccMNISTCNN-fed}
\end{table}

\subsubsection*{Results GateON, task inferred} 
\label{supp:task-inferredd}

\begin{table}[!ht]
\centering
  \begin{tabular}{|l||l|l|l|l|l|l|l|l|l|}
    \hline
    \multirow{3}{*}{Models} &
      \multicolumn{3}{c}{\textit{Permuted}} &
      \multicolumn{3}{c}{\textit{Rotated}} &
      \multicolumn{3}{c|}{\textit{Shuffled}} \\
    & $A_{cc}^{cont}$& $\Delta A^t_{cc}$ & FR. & $A_{cc}^{cont}$& $\Delta A^t_{cc}$ & FR & $A_{cc}^{cont}$ & $\Delta A^t_{cc}$ & FR \\
    \hline
    Vanilla  & 26.29 & 0.0 & 71.57 &69.68 & 0.0 & 28.53 &13.41 & 0.0 & 77.44\\
    \hline
    \hline
    % Obstruct 0& 18.69 & 34.15  & 45.82 & 48.29 & 63.85  & 29.36 & 11.15 & 13.67  & 72.16 \\
    % \hline
    % Obstruct$_{CNN}$ & 16.16 & -51.17 & 76.96 &95.07 & -1.78 & 3.73 & 15.15 & 13.67  & 72.16 \\
    % \hline
    Gating & 55.07 &  {\bf{0.24}} & 41.96 & 71.63 &  {\bf{0.94}} & 25.33 & 62.07 & {\bf{0.6}} & 35.11\\
    \hline
    \hline
    n-GateON 1& 95.75 & -0.48 & 1.12 & 97.7 & 0.7 & 0.24 &96.89 & 0.4 & 1.09 \\
    \hline
    n-GateON 0.5& 92.86 & -1.57 & 2.06 & 97.6 & 0.66 & 0.33 &97.35 & 0.39 & 0.62  \\
    \hline
    n-GateON 0& 73.64 & -5.61 & 14.82 & 97.62 & 0.59 & 0.35 &{\bf{97.72}} & 0.26 & {\bf{0.12}}  \\
    \hline
    \hline
    p-GateON 0.5& {\bf{96.48}} & -0.05 & 0.84 &97.76 & 0.55 & 0.16 &97.51 & 0.29 & 0.37 \\
    \hline
    p-GateON 0& 96.34 & -0.6 & {\bf{0.48}} &{\bf{97.87}} & 0.64 & {\bf{0.14}} &97.69 & 0.24 & 0.13  \\
    \hline
  \end{tabular}
  \caption[ $\,$ results MNIST 100, task inferred]{Performance results on the three CL-tasks (\textit{Permuted}, \textit{Rotated}, and \textit{Shuffled}) for 100 tasks with task inferred by the network.}
 \label{tabsupp:accMNIST-inferred}
\end{table}

\begin{table}[!ht]
\centering
  \begin{tabular}{|l||l|l|l|l|l|l|l|l|l|}
    \hline
    \multirow{3}{*}{CNN Models} &
      \multicolumn{3}{c}{\textit{Permuted}} &
      \multicolumn{3}{c}{\textit{Rotated}} &
      \multicolumn{3}{c|}{\textit{Shuffled}} \\
    & $A_{cc}^{cont}$& $\Delta A^t_{cc}$ & FR. & $A_{cc}^{cont}$& $\Delta A^t_{cc}$ & FR & $A_{cc}^{cont}$ & $\Delta A^t_{cc}$ & FR \\
    \hline
    Vanilla  &41.32 &   & 55.91 &94.06 &   & 5.33 &40.15 &   & 58.22 \\
    \hline
    Gating& 50.03 & {\bf{-0.07}} & 47.15 &88.76 &   -0.04  & 10.1 & 48.63 & {\bf{-0.14}} & 50.65  \\
    \hline
    n-GateON 1  &89.82 & -2.61 & 4.03 &98.92 & {\bf{0.11}} & 0.45 &98.71 & -0.16 & 0.55 \\
    \hline
    n-GateON 0.5  & 93.22 & -1.95 & {\bf{1.64}} &98.79 & 0.03 & 0.5 &98.92 & -0.18 & 0.32 \\
    \hline
    n-GateON 0 & 76.8 & -8.79 & 10.56 &98.89 & -0.01 & {\bf{0.37}} &99.02 & -0.21 & {\bf{0.2}} \\
    \hline
    \hline
    p-GateON 0.5 & {\bf{93.4}} & -1.59 & 2.51 &98.92 & 0.07 & 0.41 &98.91 & -0.15 & 0.37 \\
    \hline
    p-GateON 0 & 93.36 & -2.17 & 1.71 & {\bf{98.97}} & 0.07 & 0.36 & {\bf{99.04}} & -0.16 & 0.22 \\
    \hline
  \end{tabular}
  \caption[ $\,$ results MNIST 100, task inferred, with convolutions]{Performance results on the three CL-paradigms (\textit{Permuted}, \textit{Rotated}, and \textit{Shuffled}) for 100 tasks with task inferred by the context module of the CNN.}
 \label{tabsupp:accMNISTCNN-inferred}
\end{table}

\subsection{Time scales of freezing and unfreezing}
\label{matmet:fixed-point}
An important theoretical question is how the introduction of the threshold $\epsilon$ in equation \eqref{eq:availabilty} impacts the temporal evolution of the availability. Indeed, with $\epsilon=0$, the availability $A$ can only decrease or stay fixed. To study the evolution of $A$ for $\epsilon>0$, we consider the following toy scenario: we assume that a given neuron is relevant for the first $T$ update steps ($T$ could, for example, represent all updates during the first task) where relevance is indicated by a high level $\mu=a>\epsilon$;  and thereafter the same neuron is irrelevant for the next $T'$ steps, as indicated by $\mu=b < \epsilon$ ($T'$ could represent all updates during all other tasks). From our assumptions and Eq. \eqref{eq:availabilty} with $A(0) = A_0$ we have :
\begin{equation}
    A(T+T') = A_0 (1-\eta_A (a-\epsilon))^T (1-\eta_A (b-\epsilon))^{T'}.
\end{equation}
We ask ourselves how big $T'$ needs to be for the availability to return to its original value. Solving $A(T+T')= A_0$ we obtain
$\epsilon$ being an unstable fixed point we want to estimate if there is a possible oscillatory fixed point such that the expectation $\mathrm{E}[A]\neq 0$ or $1$. We will show with fixed $\mu$ fixed during each task that such a fixed point is almost impossible. Assume $\mu=a  >\epsilon$ for T steps and $\mu=b < \epsilon$ for T' steps. Since $\mu$s are fixed from Eq. \eqref{eq:availabilty}  for $A(0) = A_0$. we obtain:
% \begin{equation}
%     A(T+T') = A_0 (1-\eta_A (a-\epsilon))^T (1-\eta_A (b-\epsilon))^{T'}.
% \end{equation}

% For the sake of analysis, we break for the moment the logic of CL and assume that after  a time $T+T^*$, we would restart with the first task.
% If $A_0 = A(T+T^*) = A(2(T+T^*)) =...$, then the availability loops periodically. The period $T+T^*$ corresponding to a fixed point of $A(n(T+T^*))$ $=A(m(T+T^*))$ for all $n,m$ is obtained, as a function of the relevance $a$ and $b$ 
% \begin{equation}
%    (1-\eta_A (a-\epsilon))^T (1-\eta_A (b-\epsilon))^{T'} =1.
% \end{equation}
% For given relevance $a$ and $b$ and a time of training on a task $T$ the exact time during which the relevance must be equal to $b$ so we obtain a fixed point different to 0 or 1 is 
\begin{equation}
T^* = - T\frac{ \log \left(1-\eta_A(a-\epsilon)\right)}{ \log \left(1-\eta_A(b-\epsilon)\right)}.
    \label{matmet:eq:tprime}
\end{equation}
Eq. \eqref{matmet:eq:tprime} shows that the higher the relevance $a$ during the first period of $T$ steps, the longer it will take to grow back to $A_0$. For the further interpretation, we suppose now that $\epsilon=1$. Since  $0<b<\epsilon$, we have $\epsilon-b<1$ during the 'irrelevant' phase. The relevance $a$, however, is bounded by the number of neurons $N$ per layer (for n-GateON). Hence, during the 'relevant' phase of a neuron, the difference $a-\epsilon$ can take a value much larger than 1.  The behavior described by Eq. \eqref{matmet:eq:tprime}  is exactly what we expect from the threshold term. It shows that very relevant neurons ($a\gg \epsilon$) have $T^*\gg T$ meaning that, for relevant neurons, the time scale of forgetting is much slower than the time scale of learning. 
% If $T'<T^*$ then we observe that $A(k(T+T')) \rightarrow 0$  as $k\rightarrow \infty$ which stops the dynamic. If $T'>T^*$ then $A(k(T+T')) \rightarrow 1$ as $k\rightarrow \infty$  and we obtain a loop during which availability decreases during the first T steps then grows back to 1 in T'.

\end{document}